\documentclass{article}

\PassOptionsToPackage{numbers}{natbib}
\usepackage[final]{neurips_data_2021}

\usepackage[utf8]{inputenc} 
\usepackage[T1]{fontenc}    
\usepackage{url}            
\usepackage{booktabs}       
\usepackage{amsfonts}       
\usepackage{nicefrac}       
\usepackage{microtype}      
\usepackage{xcolor}         

\usepackage{graphicx}
\usepackage{float}
\usepackage{subfigure}
\usepackage{caption}
\usepackage{booktabs}
\usepackage{multirow}
\usepackage{tabularx}
\newcommand{\tabincell}[2]{\begin{tabular}{@{}#1@{}}#2\end{tabular}}
\usepackage[colorlinks,linkcolor=blue]{hyperref}
\newcommand{\ie}{\textit{i}.\textit{e}., }
\newcommand{\eg}{\textit{e}.\textit{g}. }

\newcolumntype{Z}{>{\centering\let\newline\\\arraybackslash\hspace{0pt}}X}

\title{One Million Scenes for Autonomous Driving: \\ ONCE Dataset}

\author{%
  Jiageng Mao $^1$$^*$
    
    \And
   Minzhe Niu $^2$$^*$
    
    \And
   Chenhan Jiang $^2$
     \And
   Hanxue Liang $^5$
     \And
   Jingheng Chen $^3$
     \And
   Xiaodan Liang $^4$$^\dagger$
     \And
   Yamin Li $^3$
     \And
   Chaoqiang Ye $^2$
    \And
   Wei Zhang $^2$
    \And
   Zhenguo Li $^2$
    \And
   Jie Yu $^3$
    \And
   Hang Xu $^2$$^\dagger$
    \And
   Chunjing Xu $^2$

}

\begin{document}

\maketitle

\vspace{-4mm}

\begin{abstract}


Current perception models in autonomous driving have become notorious for greatly relying on a mass of annotated data to cover unseen cases and address the long-tail problem.
On the other hand, learning from unlabeled large-scale collected data and incrementally self-training powerful recognition models have received increasing attention and may become the solutions of next-generation industry-level powerful and robust perception models in autonomous driving.
However, the research community generally suffered from data inadequacy of those essential real-world scene data, which hampers the future exploration of fully/semi/self-supervised methods for 3D perception.
In this paper, we introduce the ONCE (\textbf{O}ne millio\textbf{N} s\textbf{C}en\textbf{E}s) dataset for 3D object detection in the autonomous driving scenario. The ONCE dataset consists of $1$ million LiDAR scenes and $7$ million corresponding camera images. The data is selected from $144$ driving hours, which is 20x longer than the largest 3D autonomous driving dataset available (\eg nuScenes and Waymo), and it is collected across a range of different areas, periods and weather conditions. To facilitate future research on exploiting unlabeled data for 3D detection, we additionally provide a benchmark in which we reproduce and evaluate a variety of self-supervised and semi-supervised methods on the ONCE dataset. We conduct extensive analyses on those methods and provide valuable observations on their performance related to the scale of used data. Data, code, and more information are available at \href{https://once-for-auto-driving.github.io/index.html}{http://www.once-for-auto-driving.com}. 


\let\thefootnote\relax\footnotetext{$^*$ Equal contribution. \hspace{7mm} $^1$ The Chinese University of Hong Kong \hspace{6mm} $^2$ Huawei Noah's Ark Lab}
\let\thefootnote\relax\footnotetext{$^3$ Huawei IAS BU Vehicle Cloud Service \hspace{2mm} $^4$ Sun Yat-Sen University \hspace{5mm} $^5$ ETH Zurich}
\let\thefootnote\relax\footnotetext{$^\dagger$ Corresponding authors: \url{xu.hang@huawei.com}  \& \url{xdliang328@gmail.com}}

\end{abstract}

\section{Introduction}
Autonomous driving is a promising technology that has the potential to ease the drivers' burden and save human lives from accidents. In autonomous driving systems, 3D object detection is a crucial technique that can identify and localize the vehicles and humans surrounding the self-driving vehicle, given 3D point clouds from LiDAR sensors and 2D images from cameras as input. Recent advances~\cite{caesar2020nuscenes, sun2020scalability} in 3D object detection demonstrate that large-scale and diverse scene data can significantly improve the perception accuracy of 3D detectors.  

\begin{figure}[tp]
\centering
\includegraphics[width=1.0\textwidth]{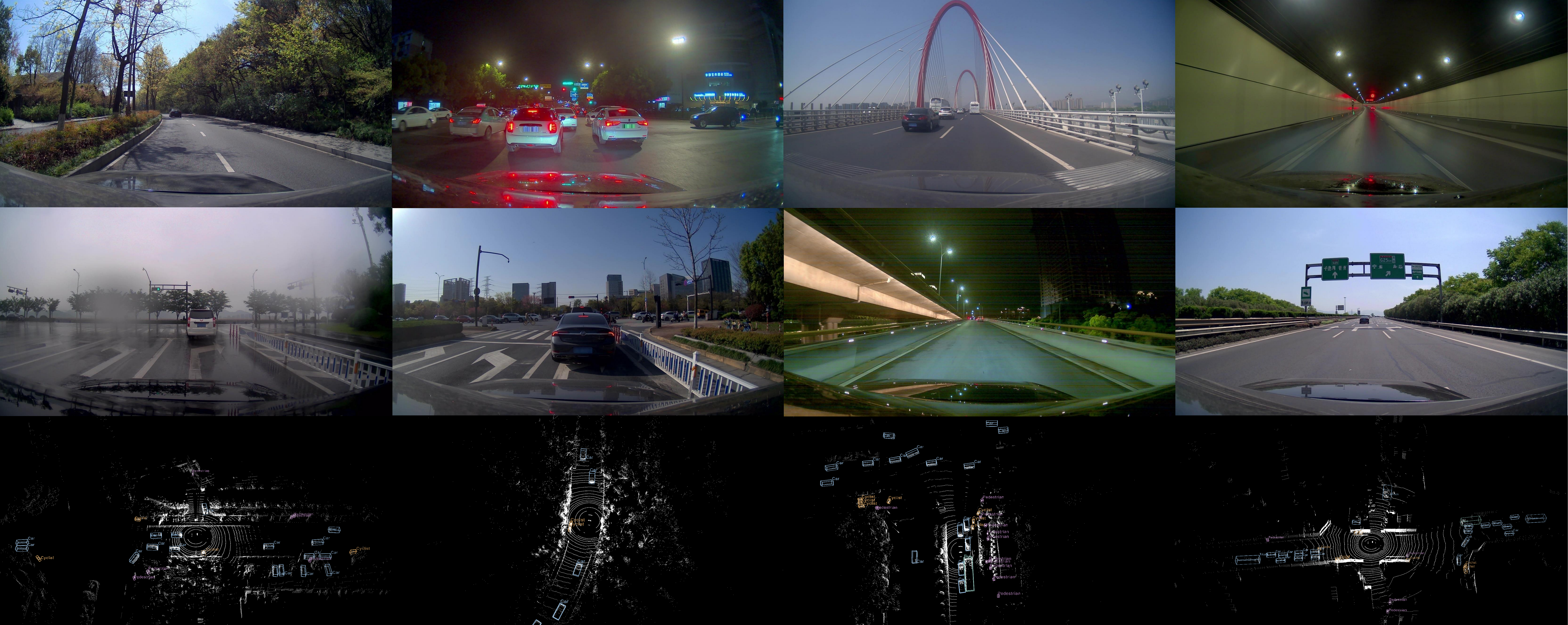}
\caption{Images and point clouds sampled from the ONCE (\textbf{O}ne millio\textbf{N} s\textbf{C}en\textbf{E}s) dataset. Our ONCE dataset covers a variety of geographical locations, time periods and weather conditions. }
\label{fig:main}
\end{figure}

Unlike other image-based datasets (\eg ImageNet~\cite{deng2009imagenet}, MS COCO~\cite{lin2014microsoft}) in which the training data can be obtained directly from websites and the annotation pipeline is relatively simple, the research community generally faces two major problems on the acquisition and exploitation of scene data for autonomous driving: 1) The data resources are scarce and the scenes generally lack diversity. The scenes for autonomous driving must be collected by driving a car that carries an expensive sensor suite on the roads in compliance with local regulations. Thus existing autonomous driving datasets could only provide a limited amount of scene data. For instance, on the largest Waymo Open dataset~\cite{sun2020scalability}, the scene data is recorded with only $6.4$ driving hours, which can hardly cover enough different circumstances. 2) Effectively leveraging unlabeled scene data becomes an important problem in practical applications. Typically a data acquisition vehicle can collect more than $200$k frames of point clouds with $8$ working hours, but a skilled worker can only annotate $100$-$200$ frames per day. This will lead to a rapid accumulation of a large amount of unlabeled data. Although algorithms of semi-supervised learning~\cite{pham2020meta, tarvainen2017mean, xie2020self}, self-supervised learning~\cite{he2020momentum, grill2020bootstrap} and unsupervised domain adaptation~\cite{long2015learning, ganin2015unsupervised} show promising results to handle those unlabeled data on the image domain, currently only a few methods~\cite{wang20203dioumatch, xie2020pointcontrast} are studied for the autonomous driving scenario, mainly because of the limited data amount provided by existing datasets.


To resolve the data inadequacy problem, in this paper, we introduce the ONCE (\textbf{O}ne millio\textbf{N} s\textbf{C}en\textbf{E}s) dataset, which is the largest and most diverse autonomous driving dataset to date. The ONCE dataset contains $1$ million 3D scenes and $7$ million corresponding 2D images, which is $5$x quantitatively more than the largest Waymo Open dataset~\cite{sun2020scalability}, and the 3D scenes are recorded with $144$ driving hours, which is $20$x longer and covers more diverse weather conditions, traffic conditions, time periods and areas than existing datasets. Figure~\ref{fig:main} shows various scenes in the ONCE dataset. Each scene is captured by a high-quality LiDAR sensor and transformed to dense 3D point clouds with $70$k points per scene in average. For each scene, $7$ cameras capture high-resolution images that cover $360^{\circ}$ field of view. The data of LiDAR sensor and cameras are precisely synchronized and additional calibration and transformation parameters are provided to enable the fusion of multiple sensors and scenes. We exhaustively annotated $16$k scenes with 3D ground truth boxes of $5$ categories (car, bus, truck, pedestrian and cyclist), giving rise to $417$k 3D boxes in total. And $769$k 2D bounding boxes are also provided for camera images by projecting 3D boxes into image planes. The other scenes are kept unannotated, mainly to facilitate future research on the exploitation of unlabeled data. Comprehensive comparisons between the ONCE dataset and other autonomous driving datasets are in Table~\ref{tab:datasets}. 

To resolve the unlabeled data exploitation problem and facilitate future research on this area, in this paper, we introduce a 3D object detection benchmark in which we implement and evaluate a variety of self-supervised and semi-supervised learning methods on the ONCE dataset. Specifically, we first carefully select a bunch of widely-used self-supervised and semi-supervised learning methods, including classic image algorithms and methods for the indoor 3D scenario. Then we adapt those methods to the task of 3D object detection for autonomous driving and reproduce their methods with the same detection framework. We train and evaluate those approaches and finally provide some observations on semi-/self-supervised learning for 3D detection by analyzing the obtained results. We also provide baseline results for multiple 3D detectors and unsupervised domain adaptation methods. Extensive experiments show that models pretrained on the ONCE dataset perform much better than those pretrained on other datasets (nuScenes and Waymo) using the same self-supervised method, which implies the superior data quality and diversity of our dataset.

Our main contributions can be summarized into two folds: 1) We introduce the ONCE dataset, which is the largest and most diverse autonomous driving dataset up to now. 2) We introduce a benchmark of self-supervised and semi-supervised learning for 3D detection in the autonomous driving scenario. 

\section{Related Work}
\textbf{Autonomous driving datasets.} Most autonomous driving datasets collect data on the roads with multiple sensors mounted on a vehicle, and the obtained point clouds and images are further annotated for perception tasks including detection and tracking. The KITTI dataset~\cite{geiger2013vision} is a pioneering work in which they record $22$ road sequences with stereo cameras and a LiDAR sensor. The ApolloScape dataset~\cite{huang2018apolloscape} offers per-pixel semantic annotations for $140$k camera images and~\cite{ma2019trafficpredict} additionally provides point cloud data based on the ApolloScape. The KAIST Multi-Spectral dataset~\cite{choi2018kaist} uses thermal imaging cameras to record scenes. The H3D dataset~\cite{patil2019h3d} provides point cloud data in $160$ urban scenes. The Argoverse dataset~\cite{chang2019argoverse} introduces geometric and semantic maps. The Lyft L5 dataset~\cite{lyft} and the A*3D dataset~\cite{pham20203d} offer $46$k and $39$k annotated LiDAR frames respectively.

\begin{table}[tp]
\setlength{\tabcolsep}{9.5pt}{
\begin{tabular}{c|ccccccc}
\toprule[2pt]
Dataset       & Scenes & \tabincell{c}{Size \\ (hr.)} & \tabincell{c}{Area \\ (km$^{2}$) } & Images & \tabincell{c}{3D \\ boxes} &  \tabincell{c}{night/\\rain} & Cls. \\
\midrule[1pt]
KITTI~\cite{geiger2013vision}   & 15k    & 1.5          & -         & 15k      & 80k     & No/No                 & 3       \\
ApolloScape~\cite{ma2019trafficpredict}   & 20k    & 2          & -                                    & 0      & 475k     & No/No                 & 6       \\
KAIST~\cite{choi2018kaist}         & 8.9k   & -          & -                                    & 8.9k   & 0        & Yes/No                & 3       \\
A2D2~\cite{geyer2020a2d2}          & 40k    & -          & -                                    & -      & -        & No/Yes                & 14      \\
H3D~\cite{patil2019h3d}           & 27k    & 0.8        & -                                    & 83k    & 1.1M     & No/No                 & 8       \\
Cityscapes 3D~\cite{gahlert2020cityscapes} & 20k    & -          & -                                    & 20k    & -        & No/No                 & 8       \\
Argoverse~\cite{chang2019argoverse}     & 44k    & 1          & 1.6                                  & 490k   & 993k     & Yes/Yes               & 15      \\
Lyft L5~\cite{lyft}       & 30k    & 2.5        & -                                    & -      & 1.3M     & No/No                 & 9       \\
A*3D~\cite{pham20203d}          & 39k    & 55         & -                                    & 39k    & 230k     & Yes/Yes               & 7       \\
nuScenes~\cite{caesar2020nuscenes}      & 400k   & 5.5        & 5                                    & 1.4M   & 1.4M     & Yes/Yes               & 23      \\
Waymo Open~\cite{sun2020scalability}    & 230k   & 6.4        & 76                                   & 1M     & 12M      & Yes/Yes               & 4       \\
\midrule[1pt]
ONCE (ours)   & \textbf{1M}     & \textbf{144}        & \textbf{210}                                  & \textbf{7M}     & 417k     & \textbf{Yes/Yes}               & 5 \\
\bottomrule[2pt]
\end{tabular}}
\vspace{2pt}
\caption{Comparisons with other 3D autonomous driving datasets. "-" means not mentioned. Our ONCE dataset has $4$x scenes, $7$x images, and $20$x driving hours compared with the largest dataset~\cite{sun2020scalability}. \label{tab:datasets}}
\end{table}

The nuScenes dataset~\cite{caesar2020nuscenes} and the Waymo Open dataset~\cite{sun2020scalability} are currently the most widely-used autonomous driving datasets. The nuScenes dataset records $5.5$ hours driving data by multiple sensors with $400$k 3D scenes in total, and the Waymo Open dataset offers $200$k scenes of $6.4$ driving hours with massive annotations. Compared with those two datasets, our ONCE dataset is not only quantitatively larger in terms of scenes and images, \eg $1$M scenes versus $200$k in~\cite{sun2020scalability}, but also more diverse since our $144$ driving hours cover all time periods as well as most weather conditions. Statistical comparisons with other autonomous driving datasets are shown in Table~\ref{tab:datasets}.

\textbf{3D object detection in driving scenarios.} Many techniques have been explored for 3D object detection in driving scenarios, and they can be broadly categorized into two classes: 1) Single-modality 3D detectors~\cite{yan2018second, shi2019pointrcnn, lang2019pointpillars, shi2020pv, yin2020center, zhou2018voxelnet, qi2018frustum, yang2018pixor} are designed to detect objects solely from sparse point clouds. PointRCNN~\cite{shi2019pointrcnn} operates directly on point clouds to predict bounding boxes. SECOND~\cite{yan2018second} rasterizes point clouds into voxels and applies 3D convolutional networks on voxel features to generate predictions. PointPillars~\cite{lang2019pointpillars} introduces the pillar representation to project point clouds to Bird Eye View (BEV) and utilizes 2D convolutional networks for object detection. PV-RCNN~\cite{shi2020pv} combines point clouds and voxels for proposal generation and refinement. CenterPoints~\cite{yin2020center} introduces the center-based assignment scheme for accurate object localization. 2) Multi-modality approaches~\cite{vora2020pointpainting, chen2017multi, ku2018joint, sindagi2019mvx, liang2019multi, yoo20203d} leverage both point clouds and images for 3D object detection. PointPainting~\cite{vora2020pointpainting} uses images to generate segmentation maps and appends the segmentation scores to corresponding point clouds to enhance point features. Other methods~\cite{chen2017multi, ku2018joint} try to fuse point features and image features on multiple stages of a detector. Our benchmark evaluates a variety of 3D detection models, including both single-modality and multi-modality detectors on the ONCE dataset. 


\textbf{Deep learning on unlabeled data.} Semi-supervised learning and self-supervised learning are two promising areas in which various emerging methods are proposed to learn from the unlabeled data effectively. Methods on semi-supervised learning are mainly of two branches: The first branch of methods try to annotate those unlabeled data with pseudo labels~\cite{lee2013pseudo, berthelot2019mixmatch, berthelot2019remixmatch, pham2020meta} by self-training~\cite{xie2020self} or teacher model~\cite{tarvainen2017mean}. Other methods~\cite{xie2019unsupervised, long2015learning, laine2016temporal, miyato2018virtual, sohn2020fixmatch} regularize pairs of augmented images under consistency constraints. Self-supervised learning approaches learn from the unlabeled data by leveraging auxiliary tasks~\cite{zhang2016colorful, noroozi2016unsupervised} or by clustering~\cite{caron2018deep, caron2020unsupervised, asano2019self}. Recent advances~\cite{he2020momentum, grill2020bootstrap, chen2020improved, chen2020simple} demonstrate that contrastive learning methods show promising results in self-supervised learning. Semi-/self-supervised learning has also been studied in 3D scenarios. SESS~\cite{zhao2020sess} is a semi-supervised method that utilizes geometric and semantic consistency for indoor 3D object detection. 3DIoUMatch~\cite{wang20203dioumatch} utilizes an auxiliary IoU head for boxes filtering. For self-supervised learning, PointContrast~\cite{xie2020pointcontrast} and DepthContrast~\cite{zhang2021self} apply contrastive learning on point clouds. Our benchmark provides a fair comparison of various self-supervised and semi-supervised methods.

\begin{figure}[tp] 
\centering 
\includegraphics[width=0.8\textwidth]{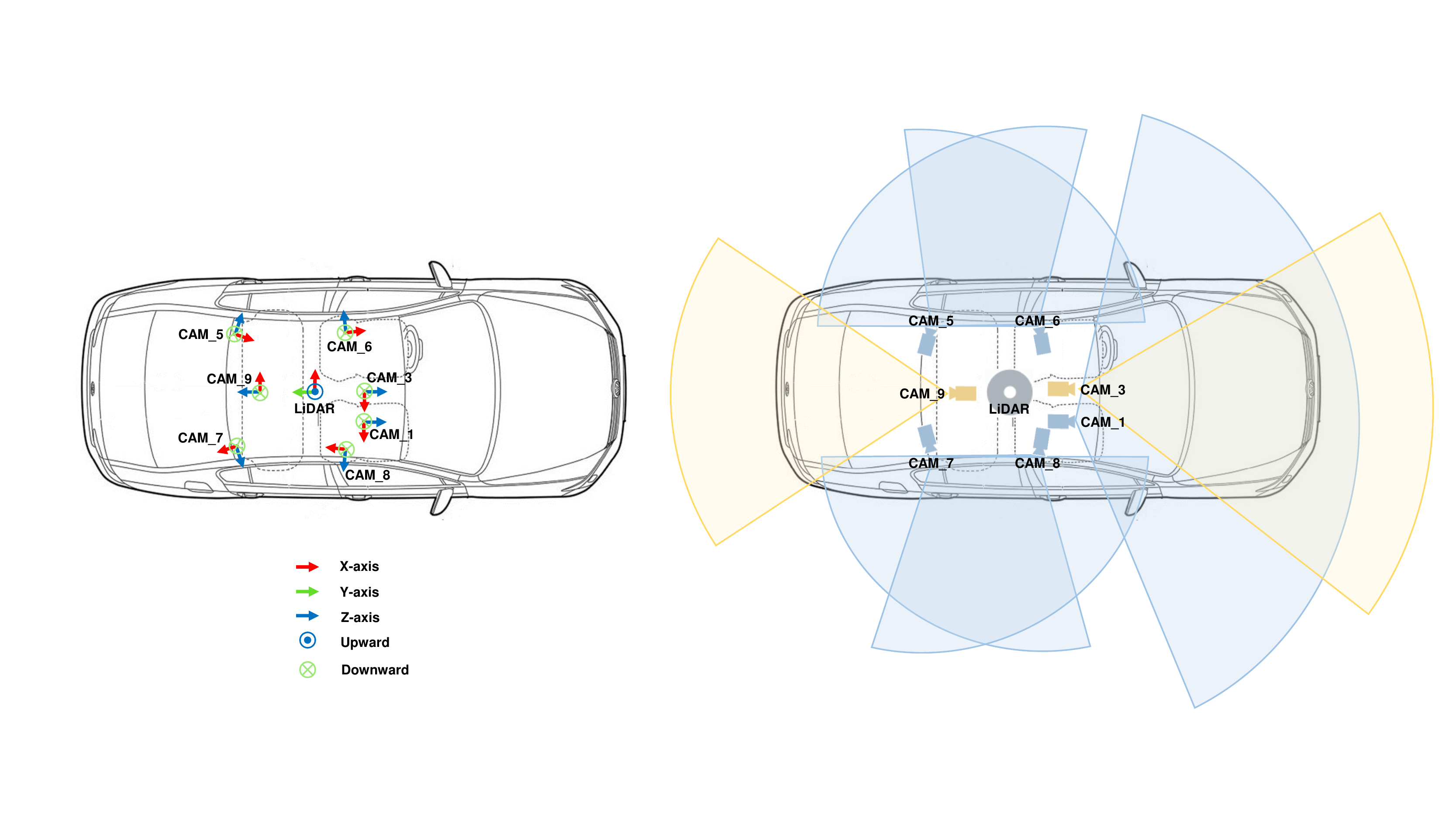}
\caption{Sensor locations and coordinate systems. The data acquisition vehicle is equipped with $1$ LiDAR and $7$ cameras that can capture 3D point clouds and images from $360^{\circ}$ field of view.}
\label{fig:sensor}
\end{figure}

\begin{table}[tp]
\resizebox{1.0\textwidth}{!}{
\begin{tabular}{c|ccccccc}
\toprule[2pt]
         & Freq. (Hz) & HFOV ($^{\circ}$) & VFOV ($^{\circ}$) & Size      & Range (m)      & Accuracy (cm) & Points/second \\
\midrule[1pt]
CAM\_1,9 & 10         & 60.6 & {[}-18, +18{]} & 1920$\times$1020 & n/a            & n/a           & n/a           \\
CAM\_3-8 & 10         & 120  & {[}-37, +37{]} & 1920$\times$1020 & n/a            & n/a           & n/a           \\
LiDAR    & 10         & 360  & {[}-25, +15{]}  & n/a       & {[}0.3, 200{]} & $\pm$2             & 7.2M
\\
\bottomrule[2pt]
\end{tabular}}
\vspace{2pt}
\caption{Detailed parameters of LiDAR and cameras.}
\label{tab:sensor}
\end{table}

\section{ONCE Dataset} \label{2}

\subsection{Data Acquisition System} \label{2.1}

\textbf{Sensor specifications.} The data acquisition system is built with one $40$-beam LiDAR sensor and seven high-resolution cameras mounted on a car. The specific sensor layout is shown in Figure~\ref{fig:sensor}, and the detailed parameters of all sensors are listed in Table~\ref{tab:sensor}. We note that the LiDAR sensor and the set of cameras can both capture data covering $360^{\circ}$ horizontal field of view near the driving vehicle, and all the sensors are well-synchronized, which enables good alignments of cross-modality data. We carefully calibrate the intrinsics and extrinsics of each camera using calibration target boards with patterns. We check the calibration parameters every day and re-calibrate the sensor that has errors. We make the intrinsics and extrinsics public along with the data to users for camera projection.

\textbf{Data collection and annotation.}  The data was collected in multiple cities in China. We conform to the local regulations and avoid releasing specific city names and locations. The data collection process lasts $3$ months. 3D ground truth boxes were manually labeled from point clouds by annotators using a commercial annotation system. The labeled boxes then went through a double-check process for validness and refinement, which guarantees high-quality bounding boxes for 3D object detection. 



\textbf{Data protection.} The driving scenes are collected in permitted areas. We comply with the local regulations and avoid releasing any localization information including GPS information and map data. For privacy protection, we actively detect any object on each image that may contain personal information, \eg human faces, license plates, with a high recall rate, and then we blur those detected objects to ensure no personal information is disclosed. 

\subsection{Data Format}


\textbf{Coordinate systems.} There are $3$ types of coordinate systems on the ONCE dataset, \ie the LiDAR coordinate, the camera coordinates, and the image coordinate. The LiDAR coordinate is placed at the center of the LiDAR sensor, with the x-axis positive to the left, the y-axis positive backwards, and the z-axis positive upwards. We additionally provide a transformation matrix (vehicle pose) between current frame and the first frame, which enables the fusion of multiple point clouds. The camera coordinates are placed at the center of the lens respectively, with the x-y plane parallel to the image plane and the z-axis positive forwards. The camera coordinates can be transformed to the LiDAR coordinate directly using the respective camera extrinsics. The image coordinate is a 2D coordinate system where the origin is at the top-left of the image, and the x-axis and the y-axis are along the image width and height respectively. The camera intrinsics enable the projection from the camera coordinate to the image coordinate. An illustration of our coordinate systems is in Figure~\ref{fig:sensor}.

\textbf{LiDAR data.} The original LiDAR data is recorded at a speed of $10$ frames per second (FPS). We further downsample those original data with the sampling rate of $2$ FPS, since most adjacent frames are quite similar thus redundant. The downsampled data is then transformed into 3D point clouds, resulting in $1$ million point clouds, \ie scenes in total. Each point cloud is represented as an $N\times4$ matrix, where $N$ is the number of points in this scene, and each point is a $4$-dim vector (x, y, z, r). The 3D coordinate (x, y, z) is based on the LiDAR coordinate, and r denotes the reflection intensity. The point clouds are stored into separate binary files for each scene and can be easily read by users.  

\textbf{Camera data.} The camera data is also downsampled along with the LiDAR data for synchronization, and then the distortions are removed to enhance the quality of the images. We finally provide JPEG compressed images for all the cameras, resulting in $7$ million images in total. 

\textbf{Annotation format.} We select $16$k most representative scenes and exhaustively annotate all the 3D bounding boxes of $5$ categories: car, bus, truck, pedestrian and cyclist. Each bounding box is a 3D cuboid and can be represented as a $7$-dim vector: (cx, cy, cz, l, w, h, $\theta$), where (cx, cy, cz) is the center of the cuboid on the LiDAR coordinate, (l, w, h) denotes length, width, height, and $\theta$ is the yaw angle of the cuboid. We provide 2D bounding boxes by projecting 3D boxes on image planes.

\textbf{Other information.} Weather and time information of each scene is useful since it contains explicit domain knowledge, but existing datasets seldom release those important data. In the ONCE dataset, we provide $3$ weather conditions, \ie sunny, cloudy, rainy, and $4$ time periods, \ie morning, noon, afternoon, night, for every labeled and unlabeled scene. We pack all the information, \ie weather, period, timestamp, pose, calibration, annotations, into a single JSON file for each scene.

\textbf{Dataset splits.} The ONCE dataset contains $581$ sequences in total. We carefully select and annotate $6$ sequences ($5$k scenes) captured in sunny days as the training split, $4$ sequences ($3$k scenes in total. $1$ sequence collected in a sunny day, $1$ in a rainy day, $1$ in a sunny night and $1$ in a rainy night) as the validation split, $10$ sequences ($8$k scenes in total. $3$ sequences in sunny days, $3$ in rainy days, $2$ in sunny nights and $2$ in rainy nights) as the testing split. We note that the sequences in each split can cover both downtown and suburban areas. The validation and testing split have quite similar data distributions, and the training split has a slight domain shift compared to the validation/testing split. We choose this way mainly to encourage the proposed methods to have better generalizability.

One major goal of the ONCE dataset is to facilitate research on leveraging large-scale unlabeled data. Thus we keep the remaining $560$ sequences as unlabeled, and those sequences can be used for semi-supervised and self-supervised 3D object detection. To explore the effects of different data amounts used for 3D detection, we also divide the unlabeled scenes into $3$ subsets: $U_{small}$, $U_{medium}$ and $U_{large}$. The small unlabeled set $U_{small}$ contains $70$ sequences ($100$k scenes), the medium set $U_{medium}$ contains $321$ sequences ($500$k scenes) and the large set $U_{large}$ contains $560$ sequences (about $1$M scenes) in total. We note that $U_{small} \subset U_{medium} \subset U_{large}$ and $U_{small}$, $U_{medium}$ are created by selecting particular roads in time order instead of uniformly sampling from all the scenes, which is more practical since the driving data is usually incrementally updated in real applications.



\subsection{Dataset Analysis}

\begin{table}[tp]
\setlength{\tabcolsep}{3.3mm}{
\begin{tabular}{c|ccc}
     \toprule[1.2pt]
     \multirow{1}{*}{{pretrain / downstream}} & {KITTI (moderate mAP)} & {nuScenes (NDS)} & {Waymo (L2 mAP)}\tabularnewline
     \midrule

     {nuScenes$\rightarrow$} & 66.1  &  -  & 53.9 \tabularnewline
    
     {Waymo$\rightarrow$} &  66.5 & 49.9 & - \tabularnewline
    
     {ONCE$\rightarrow$} &  \textbf{67.2} & \textbf{51.5}  &  \textbf{54.4} \tabularnewline
     \bottomrule[1.2pt]
\end{tabular}{\scriptsize\par}}
\vspace{1mm}
\caption{Quality analysis. The model pretrained on the ONCE dataset shows superior performance compared to those on the nuScenes and Waymo dataset, which implies our superior data quality.}
\label{tab:quality}
\end{table}

\begin{table}[tp]
    \begin{tabularx}{\textwidth}{c|ZZZ}
        \toprule[1.2pt]
         & Time & Weather & Area \\
        \midrule
        Waymo~\cite{sun2020scalability}    & daytime (80.79\%); dawn (9.04\%); night (10.17\%) & sunny (99.40\%); \newline rainy (0.60\%)    & only city-level labels   \\
        \midrule
        nuScenes~\cite{caesar2020nuscenes} & daytime (88.32\%); \newline night (11.68\%)  & sunny (80.47\%); \newline rainy (19.53\%)   & only city-level labels   \\
        \midrule
        BDD100k~\cite{yu2020bdd100k} & daytime (52.57\%); \newline dawn (7.27\%); \newline night (39.94\%); \newline undefined (0.22\%) & clear (53.45\%); \newline overcast (12.53\%); \newline partly cloudy (7.04\%); rainy (7.27\%); \newline snowy (7.91\%); foggy (0.18\%); undefined (11.61\%) & city street (62.14\%); highway (24.89\%); residential (11.68\%); parking lot (0.53\%); tunnel (0.20\%); gas stations (0.04\%); undefined (0.52\%) \\ 
        \midrule
        ONCE     & morning (39.34\%); \newline noon (3.76\%); \newline afternoon (36.67\%); \newline night (20.24\%)   & sunny (63.8\%); \newline cloudy (30.09\%); \newline rainy (6.11\%) & downtown (34.29\%); \newline suburbs (50.98\%); tunnel (1.83\%); highway (11.87\%); bridge (1.02\%)  \\
        \bottomrule[1.2pt]
    \end{tabularx}
\vspace{1mm}
\caption{Diversity analysis. Time and weather labels on nuScenes and Waymo are extracted from scene descriptions and annotations respectively. Both two datasets only provide city-level labels instead of specific road types. Compared to other datasets, our ONCE dataset contains sufficient data captured in rainy days and at nights, and we additionally provide the label of road type for each scene.}
\label{tab:diversity}
\end{table}

\textbf{Quality analysis.} In order to evaluate the data quality and provide a fair comparison across different datasets, we propose an approach that utilizes pretrained models to imply the respective data quality. Specifically, we first pretrain $3$ same backbones of the SECOND detector~\cite{yan2018second} by the self-supervised method DeepCluster~\cite{caron2018deep} using data from nuScenes~\cite{caesar2020nuscenes}, Waymo~\cite{sun2020scalability} and ONCE respectively, and then we finetune those pretrained models on multiple downstream datasets under the same settings and report their performances. The superior model should have the best pretrained backbone, which means its corresponding pretraining dataset has the best data quality. Table~\ref{tab:quality} shows the final results. The model pretrained on the ONCE dataset attains $67.2\%$ moderate mAP on the downstream KITTI~\cite{geiger2013vision} dataset, and significantly outperforms models pretrained on the Waymo dataset ($66.5\%$) and nuScenes dataset ($66.1\%$), which implies our superior data quality compared to nuScenes and Waymo. 

\textbf{Diversity analysis.} We analyze the ratios of different weather conditions, time periods and areas in the ONCE dataset and compare them to those in other datasets in Table~\ref{tab:diversity}. Our ONCE dataset contains more rainy scenes which accounts for $20\%$ of the total scenes, compared to $10\%$ in Waymo and $12\%$ in nuScenes. The ONCE dataset also provides a sufficient amount of data captured at night with $6\%$ of the total scenes. It is worth noting that we provide explicit labels of both time, weather, and area for each scene, while Waymo and nuScenes didn't provide labels of collecting areas. The driving scenes can cover most road types including downtown, suburb, highway, bridge and tunnel.

\begin{figure}[tp] 
\centering 
\includegraphics[width=1.0\textwidth]{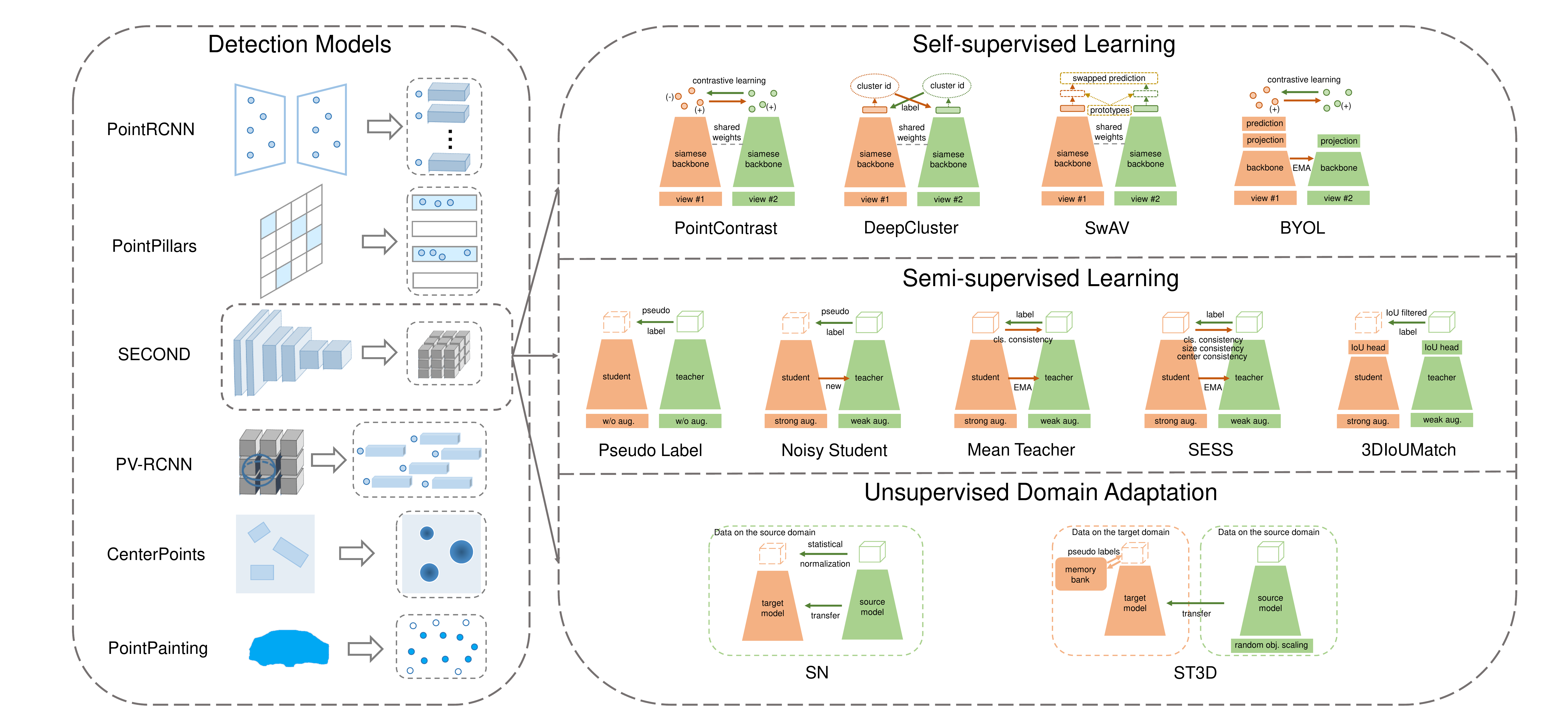}
\caption{An overview of our 3D object detection benchmark. We reproduce $6$ detection models, $4$ self-supervised learning, $5$ semi-supervised learning, and $2$ unsupervised domain adaptation methods for 3D object detection. We give comprehensive analyses on the results and offer valuable observations.}
\label{fig:overview}
\end{figure}

\section{Benchmark for 3D Object Detection} \label{4}
In this section, we present a 3D object detection benchmark on the ONCE dataset. We reproduce widely-used detection models as well as various methods of self-supervised learning, semi-supervised learning and unsupervised domain adaptation on 3D object detection. We validate those methods with a unified standard and provide performance analysis as well as suggestions for future research.  
\subsection{Models for 3D Object Detection} \label{4.1}
We implement $5$ most widely-used single-modality 3D detectors: PointRCNN~\cite{shi2019pointrcnn}, PointPillars~\cite{lang2019pointpillars}, SECOND~\cite{yan2018second}, PV-RCNN~\cite{shi2020pv} and CenterPoints~\cite{yin2020center} using only point clouds as input, as well as $1$ multi-modality detector PointPainting~\cite{vora2020pointpainting} using both point clouds and images as input on the ONCE dataset. We train those detectors on the training split and report the overall and distance-wise $AP^{Ori}_{3D}$ on the testing split using the evaluation metric in appendix E. Performance on the validation split is also reported in appendix B. We provide training and implementation details in appendix C.

\textbf{Points vs. voxels vs. pillars.} Multiple representations (points/voxels/pillars) have been explored for 3D detection. Our experiments in Table~\ref{tab:models} demonstrate that the point-based detector~\cite{shi2019pointrcnn} performs poorly with only $31.8\%$ mAP on the ONCE dataset, since small objects like pedestrians are naturally sparse and a small amount of used points cannot guarantee a high recall rate. The voxel-based detector~\cite{yan2018second} shows decent performance with $51.90\%$ mAP compared with $45.47\%$ mAP of the pillar-based detector~\cite{lang2019pointpillars}. It is mainly because voxels contain finer geometric information than pillars. PV-RCNN~\cite{shi2020pv} combines both the point representation and the voxel representation and further improves the detection performance to $53.85\%$ mAP.

\textbf{Anchor assignments vs. center assignments.} The only difference between SECOND~\cite{yan2018second} and CenterPoints~\cite{yin2020center} is that SECOND uses anchor-based target assignments while CenterPoints introduces the center-based assignments. SECOND shows better performance on the vehicle category ($69.71\%$ vs. $66.35\%$) while CenterPoints performs much better on the small objects including pedestrians ($51.80\%$ vs. $26.09\%$) and cyclists ($65.57\%$ vs. $59.92\%$) in our experiments. It is because the center-based method shows stronger localization ability which is required for detecting small objects, while the anchor-based method can estimate the size of objects more precisely.    

\textbf{Single-modality vs. multi-modality.} PointPainting~\cite{vora2020pointpainting} appends the segmentation scores to the input point clouds of CenterPoints~\cite{yin2020center}, but the performance drops from $61.24\%$ to $59.78\%$. We find that the performance of PointPainting heavily relies on the accuracy of segmentation scores, and without explicit segmentation labels on the ONCE dataset, we cannot generate accurate semantic segmentation maps from images, which brings negative effects on 3D detection.  

\begin{table}[tp]
\centering
\resizebox{\textwidth}{!}{
\begin{tabular}{c|cccc|cccc|cccc|c}
\toprule[2pt]
\multirow{2}{*}{Method} & \multicolumn{4}{|c|}{Vehicle}        & \multicolumn{4}{|c|}{Pedestrian}     & \multicolumn{4}{|c|}{Cyclist}        & \multirow{2}{*}{mAP} \\
                        & overall & 0-30m & 30-50m & 50m-inf & overall & 0-30m & 30-50m & 50m-inf & overall & 0-30m & 30-50m & 50m-inf &                      \\
\midrule[1pt]
\multicolumn{14}{c}{Multi-Modality (point clouds + images)}                                                                                                                                   \\
\midrule[1pt]
PointPainting~\cite{vora2020pointpainting}   & 66.46   & 83.70 & 56.89  & 40.74   & 47.62   & 58.95 & 39.33  & 23.34   & 65.27   & 73.48 & 61.53  & 43.90   & 59.78          \\

\midrule[1pt]
\multicolumn{14}{c}{Single-Modality (point clouds only)}                                                                                                                                   \\
\midrule[1pt]
PointRCNN~\cite{shi2019pointrcnn}               & 52.00   & 74.44 & 40.72  & 22.14   & 8.73    & 12.20 & 6.96   & 2.96    & 34.02   & 46.48 & 27.39  & 11.45   & 31.58                \\
PointPillars~\cite{lang2019pointpillars}            & 69.52   & 84.51 & 60.55  & 45.72   & 17.28   & 20.21 & 15.06  & 11.48   & 49.63   & 60.15 & 42.43  & 27.73   & 45.47                \\
SECOND~\cite{yan2018second}                  & 69.71   & 86.96 & 60.22  & 43.02   & 26.09   & 30.52 & 24.63  & 14.19   & 59.92   & 70.54 & 54.89  & 34.34   & 51.90                \\
PV-RCNN~\cite{shi2020pv}                 & 76.98   & 89.89 & 69.35  & 55.52   & 22.66   & 27.23 & 21.28  & 12.08   & 61.93   & 72.13 & 56.64  & 37.23   & 53.85                \\
CenterPoints~\cite{yin2020center}            & 66.35   & 83.65 & 56.74  & 41.57   & 51.80   & 62.80 & 45.41  & 24.53   & 65.57   & 73.02 & 62.85  & 44.77   & 61.24                \\

\bottomrule[2pt]
\end{tabular}}
\vspace{1pt}
\caption{Results of detection models on the testing split.}
\label{tab:models}
\end{table}

\subsection{Self-Supervised Learning for 3D Object Detection} \label{4.2}

We reproduce $4$ self-supervised learning methods, including $2$ contrastive learning methods (PointContrast~\cite{xie2020pointcontrast} and BYOL~\cite{grill2020bootstrap}), as well as $2$ clustering-based methods (DeepCluster~\cite{caron2018deep} and SwAV~\cite{caron2020unsupervised}) on our dataset. We choose the backbone of the SECOND detector~\cite{yan2018second} as the pretrained backbone. We first pretrain the backbone using self-supervised learning methods with different amounts of unlabeled data: $100$k $U_{small}$, $500$k $U_{medium}$ and $1$ million $U_{large}$, and then we finetune the detection model on the training split. We report the detection performances on the testing split using the evaluation metric in appendix E. Performance on the validation split is also reported in appendix B. We provide training and implementation details in appendix C.

\textbf{Self-supervised learning on unlabeled data.} Experiments in Table~\ref{tab:self} show that self-supervised methods can improve the detection results with enough unlabeled data. PointContrast~\cite{xie2020pointcontrast} obtains $50.75\%$ mAP with $100$k unlabeled data, but the performance consistently improves to $52.76\%$ and $52.99\%$ with $500$k and one million unlabeled data respectively, giving rise to $1.09\%$ final performance gain over baseline. Self-supervised learning benefits from the increasing amount of unlabeled data.

\textbf{Contrastive learning vs. clustering.} The detection results indicate that clustering-based methods~\cite{caron2018deep, caron2020unsupervised} consistently outperforms contrastive learning methods~\cite{xie2020pointcontrast, grill2020bootstrap}. SwAV~\cite{caron2020unsupervised} and DeepCluster~\cite{caron2018deep} achieve $54.28\%$ and $54.27\%$ mAP respectively on $U_{large}$, compared with $52.10\%$ and $52.99\%$ obtained by BYOL~\cite{grill2020bootstrap} and PointContrast~\cite{xie2020pointcontrast}. This is mainly because constructing representative views of a 3D scene for contrastive learning is non-trivial in driving scenarios. Generating different views simply by performing different augmentations on the same point cloud may result in quite similar views that will make the pretraining process easily converge to a trivial solution. 

\begin{table}[tp]
\resizebox{\textwidth}{!}{
\begin{tabular}{c|cccc|cccc|cccc|c}
\toprule[2pt]
\multirow{2}{*}{Method} & \multicolumn{4}{|c|}{Vehicle}        & \multicolumn{4}{|c|}{Pedestrian}     & \multicolumn{4}{|c|}{Cyclist}        & \multirow{2}{*}{mAP} \\
                        & overall & 0-30m & 30-50m & 50m-inf & overall & 0-30m & 30-50m & 50m-inf & overall & 0-30m & 30-50m & 50m-inf &                      \\
\midrule[1pt]
baseline~\cite{yan2018second}                & 69.71   & 86.96 & 60.22  & 43.02   & 26.09   & 30.52 & 24.63  & 14.19   & 59.92   & 70.54 & 54.89  & 34.34   & 51.90                 \\
\midrule[1pt]
\multicolumn{14}{c}{$U_{small}$}                                                                                                                \\
\midrule[1pt]
BYOL~\cite{grill2020bootstrap}                    & 67.57 &	84.61 &	58.26 &	41.59 &	17.22 &	19.45 &	16.71 &	10.43 &	53.36 &	64.95 &	47.47 &	27.66 &	46.05 \tiny{\textcolor[rgb]{0.6,0,0}{(-5.85)}}     \\
PointContrast~\cite{xie2020pointcontrast}            & 71.53 &	87.02 &	62.37 &	47.23 &	22.68 &	26.33 &	21.58 &	12.98 &	58.04 &	70.01 &	51.74 &	31.69 &	50.75 \tiny{\textcolor[rgb]{0.6,0,0}{(-1.15)}}	               \\
SwAV~\cite{caron2020unsupervised}            & 72.25 &	87.20 &	63.38 &	48.93 &	25.11 &	29.32 &	23.50 &	14.13 &	60.67 &	70.90 &	55.91 &	35.39 &	52.68 \tiny{\textcolor[rgb]{0,0.6,0}{(+0.78)}}               \\
DeepCluster~\cite{caron2018deep}           & 72.06 &	87.09 &	63.09 &	48.78 &	27.56 &	32.21 &	26.60 &	13.61 &	50.30 &	70.33 &	55.82 &	35.89 &	53.31 \tiny{\textcolor[rgb]{0,0.6,0}{(+1.41)}}                \\
\midrule[1pt]
\multicolumn{14}{c}{$U_{medium}$}                                                                                                                                   \\
\midrule[1pt]
BYOL~\cite{grill2020bootstrap}                    & 69.69 &	84.83 &	60.41 &	46.05 &	27.31 &	32.58 &	24.60 &	13.69 &	57.22 &	69.57 &	51.07 &	29.15 &	51.41 \tiny{\textcolor[rgb]{0.6,0,0}{(-0.49)}}                 \\
PointContrast~\cite{xie2020pointcontrast}           & 70.15 &	86.71 &	61.12 &	48.11 &	29.23 &	35.52 &	36.28 &	13.06 &	58.91 &	70.05 &	53.86 &	34.27 &	52.76 \tiny{\textcolor[rgb]{0,0.6,0}{(+0.86)}}               \\
SwAV~\cite{caron2020unsupervised}            & 72.10 &	87.11 &	63.15 &	48.58 &	28.00 &	33.10 &	25.88 &	14.19 &	60.17 &	70.46 &	55.61 &	34.84 &	53.42 \tiny{\textcolor[rgb]{0,0.6,0}{(+1.52)}}             \\
DeepCluster~\cite{caron2018deep}          & 72.12 &	87.31 &	62.97 &	48.55 &	30.06 &	36.07 &	27.23 &	13.47 &	60.45 &	70.81 &	54.93 &	36.03 &	54.21 \tiny{\textcolor[rgb]{0,0.6,0}{(+2.31)}}              \\
\midrule[1pt]
\multicolumn{14}{c}{$U_{large}$}                                                                                                                                    \\
\midrule[1pt]
BYOL~\cite{grill2020bootstrap}                    & 72.23 &	87.30 &	63.13 &	48.31 &	23.62 &	27.10 &	22.14 &	13.47 &	60.45 &	70.82 &	55.31 &	35.65 &	52.10 \tiny{\textcolor[rgb]{0,0.6,0}{(+0.20)}}                \\
PointContrast~\cite{xie2020pointcontrast}            & 73.15 &	83.92 &	67.29 &	50.97 &	27.48 &	31.45 &	24.17 &	16.70 &	58.33 &	70.37 &	52.26 &	35.61 &	52.99 \tiny{\textcolor[rgb]{0,0.6,0}{(+1.09)}}             \\
SwAV~\cite{caron2020unsupervised}            & 71.96 &	86.92 &	62.83 &	48.85 &	30.60 &	36.42 &	28.03 &	14.52 &	60.27 &	70.43 &	55.52 &	36.25 &	54.28 \tiny{\textcolor[rgb]{0,0.6,0}{(+2.38)}}                \\
DeepCluster~\cite{caron2018deep}      & 71.85 &	86.96 &	62.91 &	48.54 &	30.54 &	37.08 &	27.55 &	13.86 &	60.42 &	70.60 &	55.47 &	36.29 &	54.27 \tiny{\textcolor[rgb]{0,0.6,0}{(+2.37)}}               \\
\bottomrule[2pt]
\end{tabular}}
\vspace{1pt}
\caption{Results of self-supervised learning methods on the testing split.}
\label{tab:self}
\end{table}
\begin{table}[tp]
\resizebox{\textwidth}{!}{
\begin{tabular}{c|cccc|cccc|cccc|c}
\toprule[2pt]
\multirow{2}{*}{Method} & \multicolumn{4}{|c|}{Vehicle}        & \multicolumn{4}{|c|}{Pedestrian}     & \multicolumn{4}{|c|}{Cyclist}        & \multirow{2}{*}{mAP} \\
                        & overall & 0-30m & 30-50m & 50m-inf & overall & 0-30m & 30-50m & 50m-inf & overall & 0-30m & 30-50m & 50m-inf &                      \\
\midrule[1pt]
baseline~\cite{yan2018second}                & 69.71   & 86.96 & 60.22  & 43.02   & 26.09   & 30.52 & 24.63  & 14.19   & 59.92   & 70.54 & 54.89  & 34.34   & 51.90                 \\
\midrule[1pt]
\multicolumn{14}{c}{$U_{small}$}                                                                                                                                    \\
\midrule[1pt]
Pseudo Label~\cite{lee2013pseudo}            & 71.05   & 86.51 & 61.81  & 47.49   & 25.58   & 31.03 & 22.03  & 14.12   & 58.08   & 68.50  & 52.63  & 35.61   & 51.57 \tiny{\textcolor[rgb]{0.6,0,0}{(-0.33)}}               \\
Noisy Student~\cite{xie2020self}           & 73.25   & 88.84 & 64.61  & 49.95   & 28.04   & 34.62 & 23.43  & 14.20    & 57.58   & 67.77 & 53.43  & 33.76   & 52.96 \tiny{\textcolor[rgb]{0,0.6,0}{(+1.06)}}                \\
Mean Teacher~\cite{tarvainen2017mean}            & 74.13   & 89.34 & 65.28  & 50.91   & 31.66   & 37.44 & 29.90   & 14.61   & 62.69   & 71.88 & 59.22  & 39.45   & 56.16 \tiny{\textcolor[rgb]{0,0.6,0}{(+4.26)}}               \\
SESS~\cite{zhao2020sess}                    & 72.42   & 87.23 & 63.55  & 49.11   & 27.32   & 32.26 & 24.47  & 15.36   & 61.76   & 72.39 & 57.29  & 37.33   & 53.83 \tiny{\textcolor[rgb]{0,0.6,0}{(+1.93)}}             \\
3DIoUMatch~\cite{wang20203dioumatch}              & 72.12   & 87.05 &  63.65 &   50.35 &  31.41  &  38.56
     &   27.62    & 14.25   & 59.46  &  69.53    & 54.82  & 36.18   & 54.33 \tiny{\textcolor[rgb]{0,0.6,0}{(+2.43)}}      \\
\midrule[1pt]
\multicolumn{14}{c}{$U_{medium}$}                                                                                                                                   \\
\midrule[1pt]
Pseudo Label~\cite{lee2013pseudo}            & 70.72   & 86.21 & 61.72  & 47.39   & 21.74   & 25.73 & 19.91  & 13.28   & 56.01   & 67.14 & 50.18  & 33.23   & 49.49 \tiny{\textcolor[rgb]{0.6,0,0}{(-2.41)}}               \\
Noisy Student~\cite{xie2020self}           & 73.97   & 89.09 & 65.35  & 51.04   & 30.32   & 36.24 & 27.08  & 16.24   & 61.35   & 71.28 & 56.70   & 37.96   & 55.22 \tiny{\textcolor[rgb]{0,0.6,0}{(+3.32)}}              \\
Mean Teacher~\cite{tarvainen2017mean}            & 74.71   & 89.28 & 66.10   & 52.90    & 36.03   & 42.97 & 33.29  & 18.70    & 64.88   & 74.05 & 60.80   & 42.63   & 58.54 \tiny{\textcolor[rgb]{0,0.6,0}{(+6.64)}}             \\
SESS~\cite{zhao2020sess}                    & 72.60    & 87.02 & 64.29  & 50.68   & 35.23   & 42.59 & 31.40   & 16.64   & 64.67   & 73.93 & 61.14  & 40.80    & 57.50 \tiny{\textcolor[rgb]{0,0.6,0}{(+5.60)}}                 \\
3DIoUMatch~\cite{wang20203dioumatch}        & 74.26  &  89.08   &  66.11 &  53.03  &  33.91  & 41.02  & 30.07       &     16.15    &   61.30    & 71.29 &  56.49  &  38.13  &  56.49 \tiny{\textcolor[rgb]{0,0.6,0}{(+4.59)}}      \\
\midrule[1pt]
\multicolumn{14}{c}{$U_{large}$}                                                                                                                                    \\
\midrule[1pt]
Pseudo Label~\cite{lee2013pseudo}            & 70.29   & 85.94 & 61.18  & 46.66   & 21.85   & 25.83 & 20.22  & 12.75   & 55.72   & 66.96 & 50.29  & 32.92   & 49.29 \tiny{\textcolor[rgb]{0.6,0,0}{(-2.61)}}             \\
Noisy Student~\cite{xie2020self}           & 74.50    & 89.23 & 67.11  & 53.15   & 33.28   & 40.20  & 28.89  & 17.50    & 62.05   & 71.76 & 57.53  & 39.32   & 56.61 \tiny{\textcolor[rgb]{0,0.6,0}{(+4.71)}}               \\
Mean Teacher~\cite{tarvainen2017mean}            & 76.60    & 89.41 & 68.29  & 55.66   & 36.37   & 43.84 & 32.49  & 17.11   & 66.99   & 75.87 & 63.35  & 44.06   & 59.99 \tiny{\textcolor[rgb]{0,0.6,0}{(+8.09)}}                \\
SESS~\cite{zhao2020sess}                    & 74.52   & 88.97 & 66.32  & 52.47   & 36.29   & 43.53 & 33.15  & 16.68   & 65.52   & 74.63 & 62.67  & 41.91   & 58.78 \tiny{\textcolor[rgb]{0,0.6,0}{(+6.88)}}                \\
3DIoUMatch~\cite{wang20203dioumatch}   &  74.48  & 89.13  & 66.35  &  54.59  &  35.74   & 43.35 &  32.08
      &  17.34  &   62.06   &  71.86 & 58.00  &  39.09  & 57.43 \tiny{\textcolor[rgb]{0,0.6,0}{(+5.53)}}              \\ 
\bottomrule[2pt]
\end{tabular}}
\vspace{1pt}
\caption{Results of semi-supervised learning methods on the testing split.}
\label{tab:semi}
\end{table}

\subsection{Semi-Supervised Learning for 3D Object Detection} \label{4.3}

We implement $3$ image-based semi-supervised methods: Pseudo Label~\cite{lee2013pseudo}, Mean Teacher~\cite{tarvainen2017mean} and Noisy Student~\cite{xie2020self}, as well as $2$ semi-supervised methods for point clouds in the indoor scenario: SESS~\cite{zhao2020sess} and 3DIoUMatch~\cite{wang20203dioumatch}. We first pretrain the model on the training split and then apply the $5$ semi-supervised learning methods on both the training split and unlabeled scenes. We train those methods with $5$ epochs for the $100$k unlabeled subset $U_{small}$, $3$ epochs for the $500$k subset $U_{medium}$ and the $1$ million subset $U_{large}$ during the semi-supervised learning process. We report detection performances on the testing split with the use of unlabeled subsets $U_{small}$, $U_{medium}$ and $U_{large}$ separately. Performance on the validation split is also reported in appendix B. We provide training and implementation details in appendix C.

\textbf{Semi-supervised learning on unlabeled data.} Experiments in Table~\ref{tab:semi} show that most semi-supervised methods can improve the detection results using unlabeled data. For instance, Mean Teacher~\cite{tarvainen2017mean} improves the baseline result by $8.09\%$ in mAP using the largest unlabeled set $U_{large}$. The detection performance can be further boosted when the amount of unlabeled data increases. SESS~\cite{zhao2020sess} obtains $1.93\%$ performance gain using $100$k unlabeled scenes, and the performance gain reaches $5.60\%$ with $500$k unlabeled scenes and $6.88\%$ with one million scenes.

\textbf{Pseudo labels vs. consistency.} There are two keys to the success of labeling-based methods~\cite{lee2013pseudo, xie2020self, wang20203dioumatch}: augmentations and label quality. Without strong augmentations, the performance of Pseudo Label~\cite{lee2013pseudo} drops from $51.90\%$ to $49.29\%$ albeit one million scenes $U_{large}$ are provided for training. 3DIoUMatch~\cite{wang20203dioumatch} adds additional step to filter out labels of low quality, and the performance reaches $57.43\%$ compared with $56.61\%$ of Noisy Student on $U_{large}$. Consistency-based methods~\cite{tarvainen2017mean, zhao2020sess} generally perform better than labeling-based methods, and Mean Teacher obtains the highest performance $59.99\%$ on $U_{large}$. SESS~\cite{zhao2020sess} performs worse than Mean Teacher with $58.78\%$ mAP, which indicates that size and center consistency may not be useful in driving scenarios.    

\textbf{Semi-supervised learning vs. self-supervised learning.} Our results in Table~\ref{tab:self} and Table~\ref{tab:semi} demonstrate that the semi-supervised methods generally have a better performance compared with the self-supervised methods. Mean Teacher~\cite{tarvainen2017mean} attains the best performance of $59.99\%$ mAP while the best self-supervised method SwAV~\cite{caron2020unsupervised} only obtains $54.28\%$ on $U_{large}$. The major reason is that in semi-supervised learning the model usually receives stronger and more precise supervisory signals, \eg labels or consistency with a trained model, when learning from the unlabeled data. However, in self-supervised learning, the supervisory signals on the unlabeled data are cluster id or similarity of pairs, which are typically noisy and uncertain.


\begin{table}[tp]
\setlength{\tabcolsep}{7pt}{
\begin{tabular}{c|cc|cc|cc}
\toprule[2pt]
\multirow{2}{*}{\footnotesize{}Task} & \multicolumn{2}{|c|}{\multirow{2}{*}{\footnotesize{}Waymo $\rightarrow$ ONCE}} & \multicolumn{2}{|c|}{\multirow{2}{*}{\footnotesize{}nuScenes $\rightarrow$ ONCE}} & \multicolumn{2}{|c}{\multirow{2}{*}{\footnotesize{} ONCE $\rightarrow$ KITTI}} \\
                      & \multicolumn{2}{|c|}{}                                          & \multicolumn{2}{|c|}{}                                             & \multicolumn{2}{|c}{}                                          \\
\midrule[1pt]
\footnotesize{}Method                & \footnotesize{}$AP_{BEV}$          & \footnotesize{}$AP_{3D}$         & \footnotesize{}$AP_{BEV}$                         
& \footnotesize{}$AP_{3D}$                         & \footnotesize{}$AP_{BEV}$                        & \footnotesize{}$AP_{3D}$         \\
\midrule[1pt]
\footnotesize{}Source Only           & \footnotesize{}65.55                       & \footnotesize{}32.88                        & \footnotesize{}46.85                         & \footnotesize{}23.74                        &   \footnotesize{}42.01                 & \footnotesize{}12.11                 \\
\footnotesize{}SN~\cite{wang2020train} & \footnotesize{}67.97 \tiny{\textcolor[rgb]{0,0.6,0}{(+2.42)}}  & \footnotesize{}38.25 \tiny{\textcolor[rgb]{0,0.6,0}{(+5.67)}}          & \footnotesize{}62.47 \tiny{\textcolor[rgb]{0,0.6,0}{(+15.62)}}                & \footnotesize{}29.53 \tiny{\textcolor[rgb]{0,0.6,0}{(+5.79)}}           &   \footnotesize{}48.12 \tiny{\textcolor[rgb]{0,0.6,0}{(+6.11)}}      &   \footnotesize{}21.12 \tiny{\textcolor[rgb]{0,0.6,0}{(+9.01)}}      \\
\footnotesize{}ST3D~\cite{yang2021st3d} & \footnotesize{}68.05 \tiny{\textcolor[rgb]{0,0.6,0}{(+2.50)}}     & \footnotesize{}48.34 \tiny{\textcolor[rgb]{0,0.6,0}{(+15.46)}}     & \footnotesize{}42.53 \tiny{\textcolor[rgb]{0.6,0,0}{(-4.32)}}                        & \footnotesize{}17.52 \tiny{\textcolor[rgb]{0.6,0,0}{(-6.22)}}       &    \footnotesize{}86.89 \tiny{\textcolor[rgb]{0,0.6,0}{(+44.88)}}     &   \footnotesize{}41.42 \tiny{\textcolor[rgb]{0,0.6,0}{(+29.31)}}   \\
\footnotesize{}Oracle                &   \footnotesize{}89.00         &  \footnotesize{}77.50     &  \footnotesize{}89.00       &  \footnotesize{}77.50      &      \footnotesize{}83.29            & \footnotesize{}73.45   \\
\bottomrule[2pt]
\end{tabular}}
\vspace{1pt}
\caption{Results on unsupervised domain adaptation. Source Only means trained on the source and directly evaluated on the target domain. Oracle means trained and tested both on the target domain.}
\label{tab:uda}
\end{table}

\subsection{Unsupervised Domain Adaptation for 3D Object Detection} \label{4.4}
Unsupervised domain adaptation for 3D object detection aims to adapt a detection model from the source dataset to the target dataset without supervisory signals on the target domain. Different datasets typically have different collected environments, sensor locations and point cloud densities. In this paper, we reproduce $2$ commonly-used methods: SN~\cite{wang2020train} and ST3D~\cite{yang2021st3d}. We follow the settings of those methods and conduct experiments on transferring the model trained on the nuScenes and Waymo Open dataset to our ONCE dataset, as well as transferring the model trained on the ONCE dataset to the KITTI dataset. The detection results of the car class are reported on the respective target validation or testing set using the KITTI $AP_{3D}$ metric following~\cite{wang2020train, yang2021st3d}.

\textbf{Statistical normalization vs. self-training.} The normalization-based method SN~\cite{wang2020train} surpasses the Source Only model by $2.42\%$ in $AP_{BEV}$ and $5.67\%$ in $AP_{3D}$ on the Waymo $\rightarrow$ ONCE adaptation task, and the self-training method ST3D~\cite{yang2021st3d} also attains a considerable performance gain with $15.46\%$ $AP_{3D}$ improvement. However, ST3D performs poorly on the nuScenes $\rightarrow$ ONCE task. It is mainly because the nuScenes dataset has fewer LiDAR beams, and the model trained on nuScenes may produce more low-quality pseudo labels, which will harm the self-training process of ST3D. Although those two methods achieve strong results on the adaptation from/to the ONCE dataset, there is still a gap with the Oracle results, leaving large space for future research.
\section{Conclusion}
In this paper, we introduce the ONCE (\textbf{O}ne millio\textbf{N} s\textbf{C}en\textbf{E}s) dataset, which is the largest autonomous driving dataset to date. To facilitate future research on 3D object detection, we additionally provide a benchmark for detection models and methods of self-supervised learning, semi-supervised learning and unsupervised domain adaptation. For future works, we plan to support more tasks on autonomous driving, including 2D object detection, 3D semantic segmentation and planning.

\clearpage

{
\small

\bibliographystyle{plainnat}
\bibliography{main}

}

\clearpage

\appendix

\section{The ONCE dataset}

We publish the ONCE dataset, benchmark, develop kit, data format
and annotation instructions at our website \href{https://once-for-auto-driving.github.io/index.html}{http://www.once-for-auto-driving.com}. It is our priority to protect
the privacy of third parties. We bear all responsibility in case of
violation of rights, etc., and confirmation of the data license. 


\textbf{Dataset documentation.} \href{https://once-for-auto-driving.github.io/documentation.html}{http://www.once-for-auto-driving.com/documentation.html} shows the dataset documentation and intended uses.

\textbf{Terms of use, privacy and License.} The ONCE dataset is published under CC BY-NC-SA
4.0 license, which means everyone can use this dataset for non-commercial research
purpose. The detailed Terms of use, privacy terms and license are in \href{https://once-for-auto-driving.github.io/terms_of_use.html}{http://www.once-for-auto-driving.com/terms\_of\_use.html}.

\textbf{Data maintenance.} \href{https://once-for-auto-driving.github.io/download.html}{http://www.once-for-auto-driving.com/download.html} provides data download links for users. Data is stored in Google Drive for global users, and another copy of data is stored in BaiduYunPan for Chinese users. We will maintain the data for a long time and check the data accessibility in a regular basis.

\textbf{Benchmark and code.} \href{https://once-for-auto-driving.github.io/benchmark.html}{http://www.once-for-auto-driving.com/benchmark.html} provides benchmark results. The reproduction code will be released upon acceptance. 

\begin{figure*}[h]
\centering
\subfigure[Weather conditions]{
\includegraphics[width=0.3\textwidth]{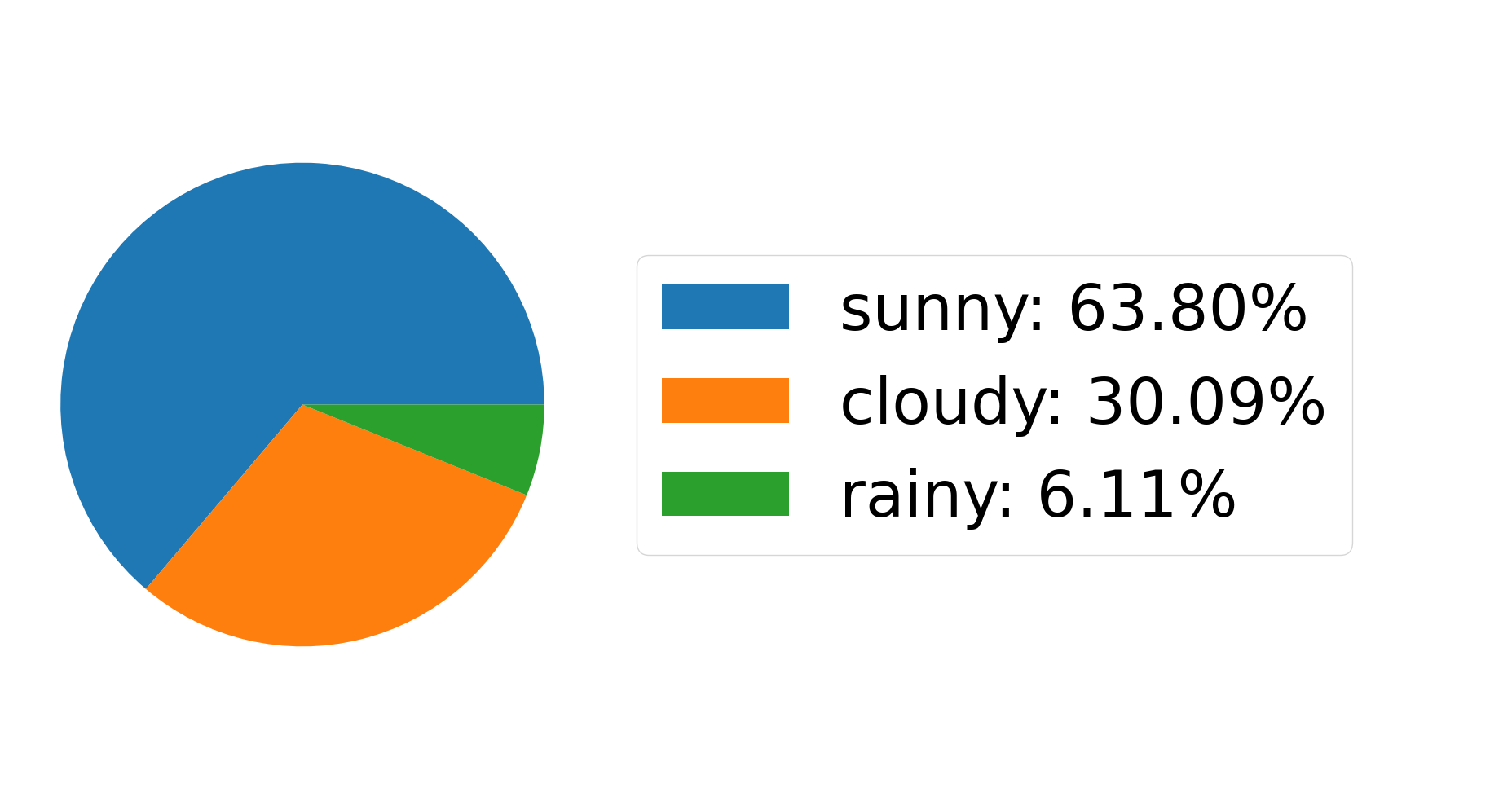}
}
\subfigure[Time periods]{
\includegraphics[width=0.3\textwidth]{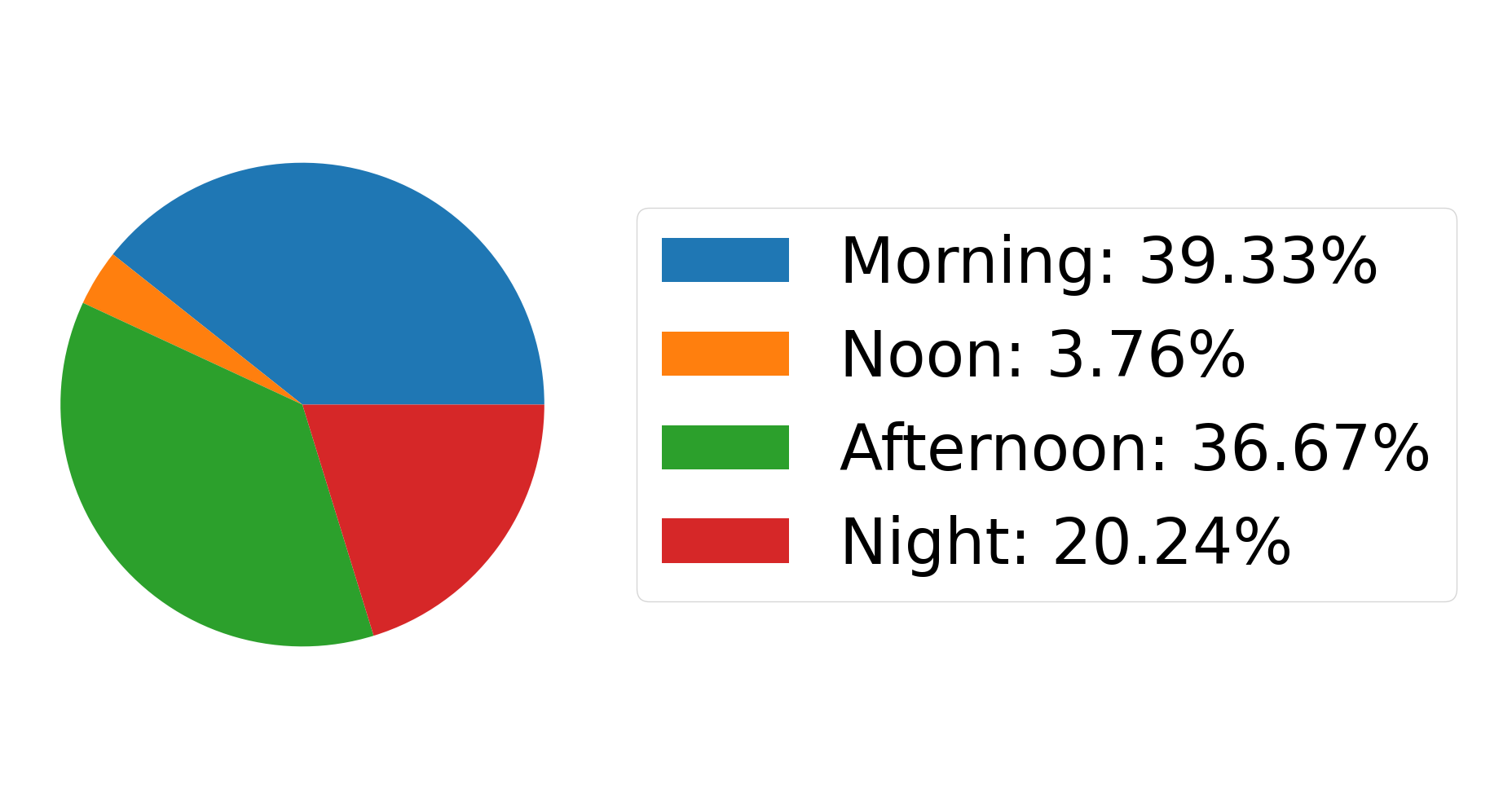}
}
\subfigure[Areas]{
\includegraphics[width=0.3\textwidth]{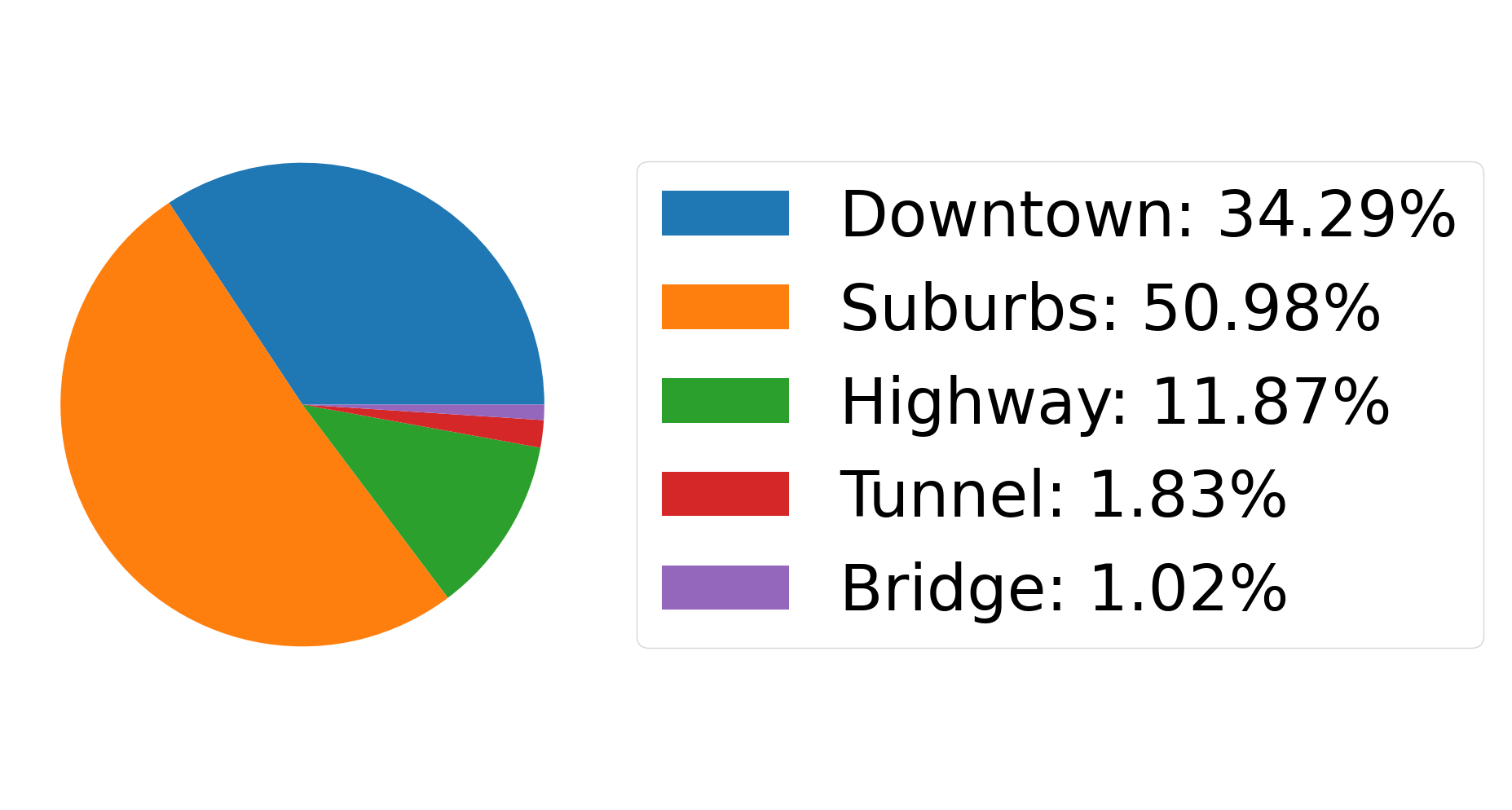}
}
\caption{Proportions of different weather, time and areas in the ONCE dataset. Our dataset covers a wide range of domains with $6\%$ scenes captured on rainy days and $20\%$ scenes collected at night. }
\label{fig:analysis_1}
\end{figure*}

\textbf{Data statistics.} Figure~\ref{fig:analysis_1} shows the proportions of different weather conditions, time periods and areas in the ONCE dataset.

\begin{figure*}[h]
\centering
\subfigure[Vehicle]{
\includegraphics[width=0.3\textwidth]{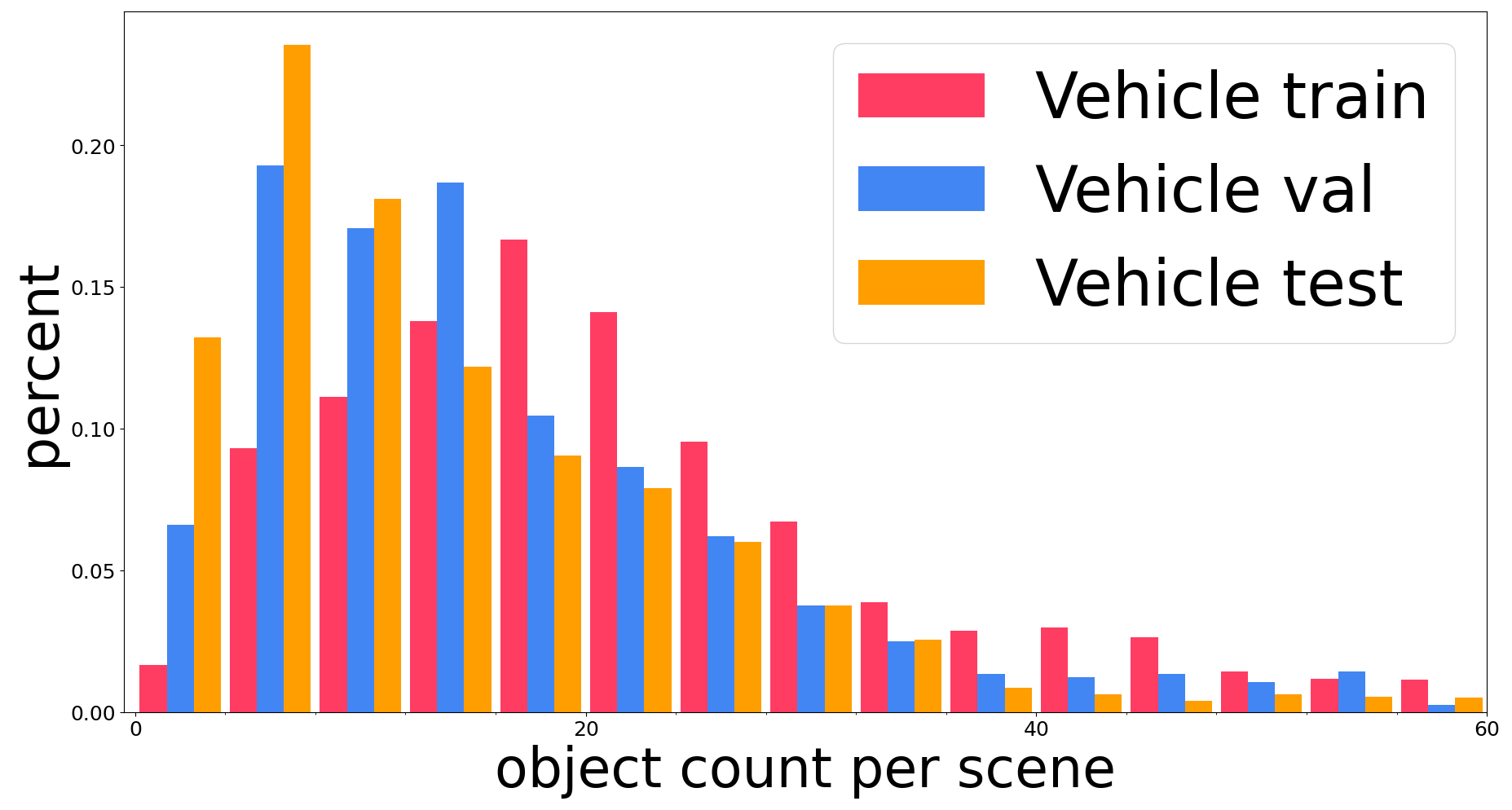}
}
\subfigure[Pedestrian]{
\includegraphics[width=0.3\textwidth]{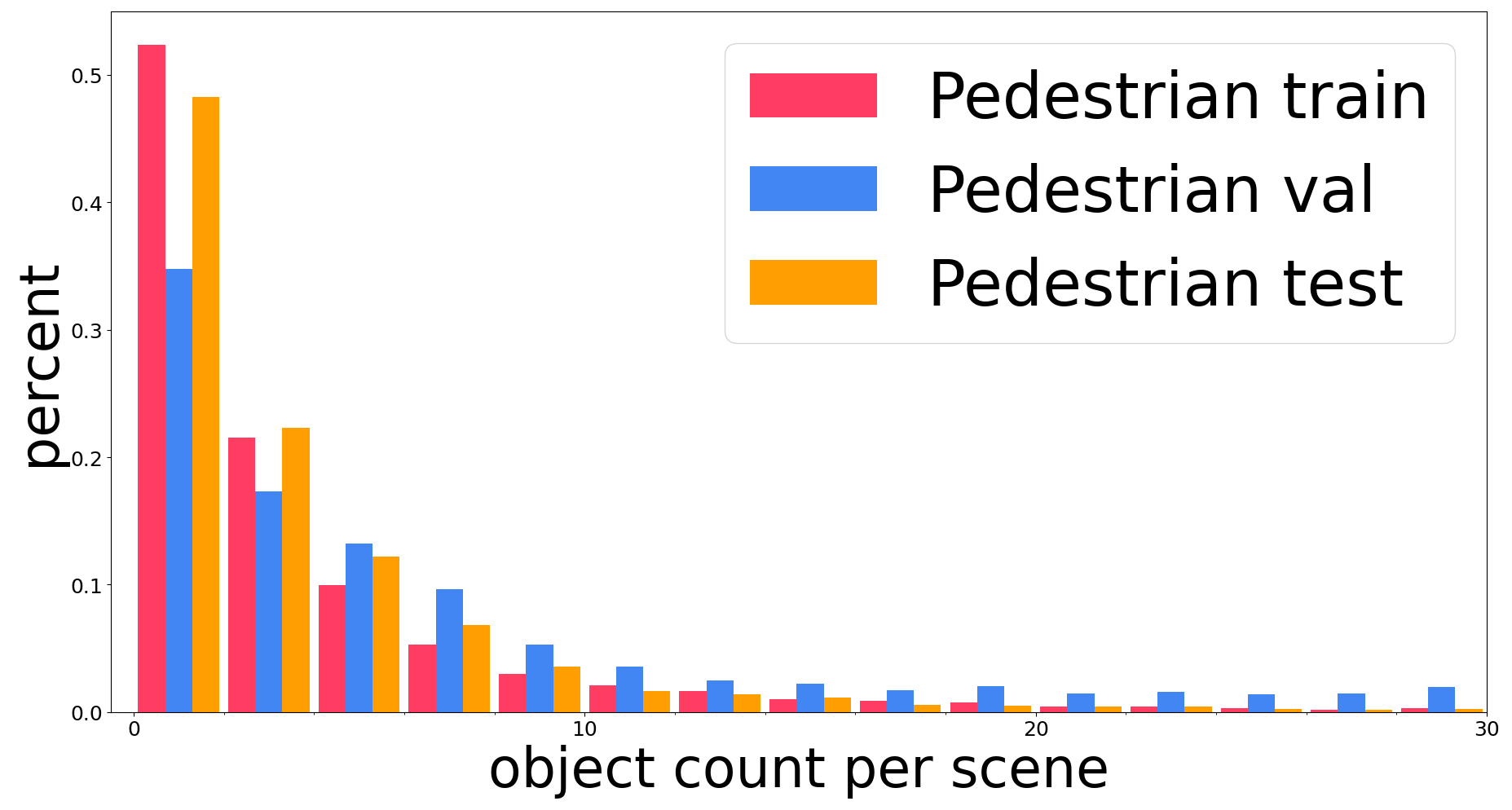}
}
\subfigure[Cyclist]{
\includegraphics[width=0.3\textwidth]{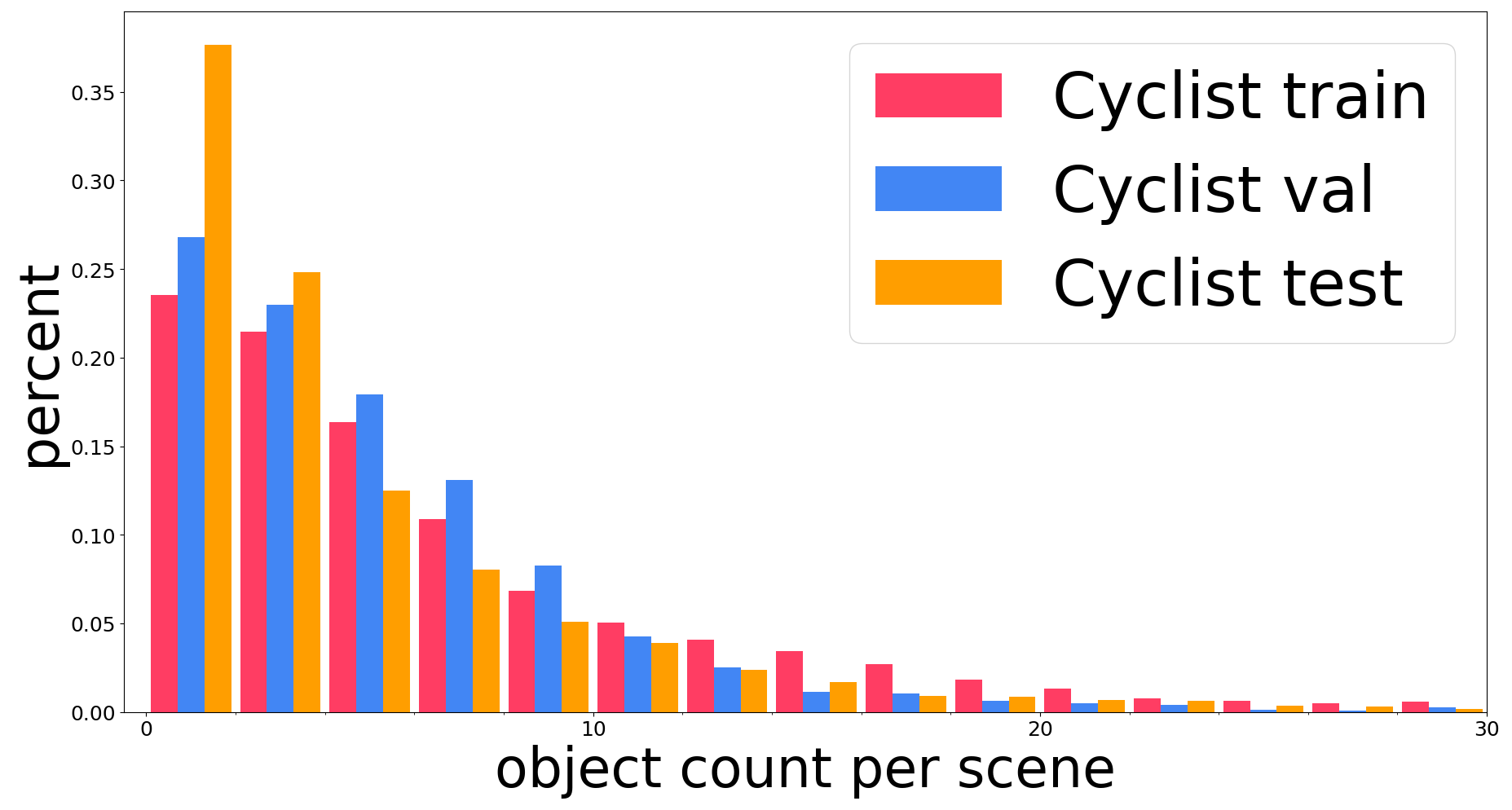}
}
\caption{Distribution of annotation counts per scene. Our ONCE dataset is diverse in the number of objects in each scene. The vehicle count in each scene ranges from $0$ to $60$.}
\label{fig:analysis_2}
\vspace{-5mm}
\end{figure*}

\textbf{Annotation statistics.} Figure~\ref{fig:analysis_2} shows the distribution of the number of objects for annotated scenes. Our annotated set covers diverse object counts for different scenes, \eg the vehicle count in a scene ranges from less than $1$ to more than $50$. Pedestrians and cyclists in a scene range from less than $1$ to more than $30$. The distributions of training, validation and testing splits are mostly similar but slightly different in some intervals, which guarantees stable evaluation results and encourages evaluated methods to have stronger generalizability across the three splits.  

\textbf{Limitations.} The major limitation of our ONCE dataset is that currently we only annotate a small amount of scenes of the one million scenes, which may hamper the broader exploration on 3D object detection. To overcome the limitation, we plan to provide more annotations in the near future. We also plan to support more autonomous driving tasks in addition to 3D detection on the ONCE dataset.  

\section{Experiments}

\subsection{Models for 3D Object Detection}
\begin{table}[H]
\centering
\resizebox{\textwidth}{!}{
\begin{tabular}{c|cccc|cccc|cccc|c}
\toprule[2pt]
\multirow{2}{*}{Method} & \multicolumn{4}{|c|}{Vehicle}        & \multicolumn{4}{|c|}{Pedestrian}     & \multicolumn{4}{|c|}{Cyclist}        & \multirow{2}{*}{mAP} \\
                        & overall & 0-30m & 30-50m & 50m-inf & overall & 0-30m & 30-50m & 50m-inf & overall & 0-30m & 30-50m & 50m-inf &                      \\
\midrule[1pt]
\multicolumn{14}{c}{Multi-Modality (point clouds + images)}                                                                                                                                   \\
\midrule[1pt]
PointPainting~\cite{vora2020pointpainting}   & 66.17 & 80.31	& 59.80	& 42.26	& 44.84	& 52.63	& 36.63	& 22.47 &	62.34 &	73.55 &	57.20 &	40.39 &	57.78          \\

\midrule[1pt]
\multicolumn{14}{c}{Single-Modality (point clouds only)}                                                                                                                                   \\
\midrule[1pt]
PointRCNN~\cite{shi2019pointrcnn}               & 52.09	& 74.45 &	40.89 &	16.81 &	4.28 &	6.17	& 2.40 &	0.91	& 29.84 &	46.03 &	20.94 &	5.46 &	28.74                \\
PointPillars~\cite{lang2019pointpillars}            & 68.57 &	80.86 &	62.07 &	47.04 &	17.63 &	19.74 &	15.15	& 10.23 &	46.81 &	58.33 &	40.32 &	25.86 &	44.34                \\
SECOND~\cite{yan2018second}                  & 71.19 &	84.04 &	63.02 &	47.25 &	26.44 &	29.33 &	24.05 &	18.05 &	58.04 &	69.96 &	52.43 &	34.61 &	51.89                \\
PV-RCNN~\cite{shi2020pv}                 & 77.77 &	89.39 &	72.55 &	58.64 &	23.50 &	25.61 &	22.84 &	17.27 &	59.37 &	71.66 &	52.58 &	36.17 &	53.55                \\
CenterPoints~\cite{yin2020center}            & 66.79 &	80.10 &	59.55 &	43.39 &	49.90 &	56.24 &	42.61 &	26.27 &	63.45 &	74.28 &	57.94 &	41.48 &	60.05                \\

\bottomrule[2pt]
\end{tabular}}
\vspace{1pt}
\caption{Results of detection models on the validation split.}
\end{table}

\subsection{Self-Supervised Learning for 3D Object Detection}
\begin{table}[H]
\resizebox{\textwidth}{!}{
\begin{tabular}{c|cccc|cccc|cccc|c}
\toprule[2pt]
\multirow{2}{*}{Method} & \multicolumn{4}{|c|}{Vehicle}        & \multicolumn{4}{|c|}{Pedestrian}     & \multicolumn{4}{|c|}{Cyclist}        & \multirow{2}{*}{mAP} \\
                        & overall & 0-30m & 30-50m & 50m-inf & overall & 0-30m & 30-50m & 50m-inf & overall & 0-30m & 30-50m & 50m-inf &                      \\
\midrule[1pt]
baseline~\cite{yan2018second}                & 71.19 &	84.04 &	63.02 &	47.25 &	26.44 &	29.33 &	24.05 &	18.05 & 58.04 &	69.96 &	52.43 &	34.61 &	51.89                 \\
\midrule[1pt]
\multicolumn{14}{c}{$U_{small}$}                                                                                                                \\
\midrule[1pt]
BYOL~\cite{grill2020bootstrap}                    & 68.02 &	81.01 &	60.21 &	44.17 &	19.50 &	22.16 &	16.68 &	12.06 &	50.61 &	62.46 &	44.29 &	28.18 &	46.04 \tiny{(-5.85)}     \\
PointContrast~\cite{xie2020pointcontrast}            & 71.07 &	83.31 &	64.90 &	49.34 &	22.52 &	23.73 &	21.81 &	16.06 &	56.36 &	68.11 &	50.35 &	34.06 &	49.98 \tiny{(-1.91)}	               \\
SwAV~\cite{caron2020unsupervised}            & 72.71 &	83.68 &	65.91 &	50.10 &	25.13 &	27.77 &	22.77 &	16.36 &	58.05 &	69.99 &	52.23 &	34.86 &	51.96 \tiny{(+0.07)}               \\
DeepCluster~\cite{tian2017deepcluster}           & 73.19 &	84.25 &	66.86 &	50.47 &	24.00 &	26.36 &	21.73 &	16.79 &	58.99 &	70.80 &	53.66 &	36.17 &	52.06 \tiny{(+0.17)}                \\
\midrule[1pt]
\multicolumn{14}{c}{$U_{medium}$}                                                                                                                                   \\
\midrule[1pt]
BYOL~\cite{grill2020bootstrap}                    & 70.93 &	84.15 &	63.48 &	45.74 &	25.86 &	29.91 &	21.55 &	15.83 &	55.63 &	58.59 &	49.01 &	29.53 &	50.82 \tiny{(-1.07)}                 \\
PointContrast~\cite{xie2020pointcontrast}           & 71.39 &	83.89 &	65.22 &	47.73 &	27.69 &	32.53 &	23.00&	14.68 &	56.88 &	69.01 &	50.41 &	34.57 &	51.99 \tiny{(+0.10)}               \\
SwAV~\cite{caron2020unsupervised}            & 72.51 &	83.39 &	65.46 &	51.08 &	27.08 &	29.94 &	25.19 &	17.13 &	57.85 &	69.87 &	52.38 &	33.78 &	52.48 \tiny{(+0.59)}             \\
DeepCluster~\cite{tian2017deepcluster}          & 71.62 &	83.99 &	65.55 &	50.77 &	29.33 &	33.25 &	25.08 &	17.00 &	57.61 &	68.57 &	52.58 &	34.05 &	52.86 \tiny{(+0.97)}              \\
\midrule[1pt]
\multicolumn{14}{c}{$U_{large}$}                                                                                                                                    \\
\midrule[1pt]
BYOL~\cite{grill2020bootstrap}                    & 71.32 &	83.59 &	64.89 &	50.27 &	25.02 &	27.06 &	22.96 &	17.04 &	58.56 &	70.18 &	52.74 &	36.32 &	51.63 \tiny{(-0.26)}                \\
PointContrast~\cite{xie2020pointcontrast}            & 71.87 &	86.93 &	62.85 &	48.65 &	28.03 &	33.07 &	25.91 &	14.44 &	60.88 &	71.12 &	55.77 &	36.78 &	53.59 \tiny{(+1.70)}             \\
SwAV~\cite{caron2020unsupervised}            & 72.46 &	83.09 &	66.66 &	51.50 &	29.84 &	34.15 &	26.22 &	17.61 &	57.84 &	68.79 &	52.21 &	35.39 &	53.38 \tiny{(+1.49)}                \\
DeepCluster~\cite{tian2017deepcluster}      & 72.89 &	83.52 &	67.09 &	50.38 &	30.32 &	34.76 &	26.43 &	18.33&	57.94 &	69.18 &	52.42 &	34.36 &	53.72 \tiny{(+1.83)}               \\
\bottomrule[2pt]
\end{tabular}}
\vspace{1pt}
\caption{Results of self-supervised learning methods on the validation split.}
\end{table}

\subsection{Semi-Supervised Learning for 3D Object Detection}

\begin{table}[H]
\resizebox{\textwidth}{!}{
\begin{tabular}{c|cccc|cccc|cccc|c}
\toprule[2pt]
\multirow{2}{*}{Method} & \multicolumn{4}{|c|}{Vehicle}        & \multicolumn{4}{|c|}{Pedestrian}     & \multicolumn{4}{|c|}{Cyclist}        & \multirow{2}{*}{mAP} \\
                        & overall & 0-30m & 30-50m & 50m-inf & overall & 0-30m & 30-50m & 50m-inf & overall & 0-30m & 30-50m & 50m-inf &                      \\
\midrule[1pt]
baseline~\cite{yan2018second}                & 71.19 &	84.04 &	63.02 &	47.25 &	26.44 &	29.33 &	24.05 &	18.05 & 58.04 &	69.96 &	52.43 &	34.61 &	51.89                 \\
\midrule[1pt]
\multicolumn{14}{c}{$U_{small}$}                                                                                                                                    \\
\midrule[1pt]
Pseudo Label~\cite{lee2013pseudo}            & 72.80 &	84.46 &	64.97 &	51.46 &	25.50 &	28.36 &	22.66 &	18.51 &	55.37 &	65.95 &	50.34 &	34.42 &	51.22 \tiny{(-0.67)}               \\
Noisy Student~\cite{xie2020self}           & 73.69 &	84.69 &	67.72 &	53.41 &	28.81 &	33.23 &	23.42 &	16.93 &	54.67 &	65.58 &	50.43 &	32.65 &	52.39 \tiny{(+0.50)}                \\
Mean Teacher~\cite{tarvainen2017mean}            & 74.46 &	86.65 &	68.44 &	53.59 &	30.54 &	34.24 &	26.31 &	20.12 &	61.02 &	72.51 &	55.24 &	39.11 &	55.34 \tiny{(+3.45)}               \\
SESS~\cite{zhao2020sess}                    & 73.33 &	84.52 &	66.22 &	52.83 &	27.31 &	31.11 &	23.94 &	19.01 &	59.52 &	71.03 &	53.93 &	36.68 &	53.39 \tiny{(+1.50)}             \\
3DIoUMatch~\cite{wang20203dioumatch}              & 73.81 &	84.61 &	68.11 &	54.48 &	30.86 &	35.87 &	25.55 &	18.30 &	56.77 &	68.02 &	51.80 &	35.91 &	53.81 \tiny{(+1.92)}      \\
\midrule[1pt]
\multicolumn{14}{c}{$U_{medium}$}                                                                                                                                   \\
\midrule[1pt]
Pseudo Label~\cite{lee2013pseudo}            & 73.03 &	86.06 &	65.96 &	51.42 &	24.56 &	27.28 &	20.81 &	17.00 &	53.61 &	65.26 &	48.44 &	33.58 &	50.40 \tiny{(-1.49)}               \\
Noisy Student~\cite{xie2020self}           & 75.53 &	86.52 &	69.78 &	55.05 &	31.56 &	35.80 &	26.24 &	21.21	& 58.93 &	69.61 &	53.73 &	36.94 &	55.34 \tiny{(+3.45)}              \\
Mean Teacher~\cite{tarvainen2017mean}            & 76.01 &	86.47 &	70.34 &	55.92 &	35.58 &	40.86 &	30.44 &	19.82 &	63.21 &	74.89 &	56.77 &	40.29 &	58.27 \tiny{(+6.38)}             \\
SESS~\cite{zhao2020sess}                    & 72.11 &	84.06 &	66.44 &	53.61 &	33.44 &	38.58 &	28.10 &	18.67 &	61.82 &	73.20 &	56.60 &	38.73 &	55.79 \tiny{(+3.90)}                 \\
3DIoUMatch~\cite{wang20203dioumatch}        & 75.69 &	86.46 &	70.22 &	56.06 &	34.14 &	38.84 &	29.19 &	19.62 &	58.93 &	69.08 &	54.16 &	38.87 &	56.25 \tiny{(+4.36)}      \\
\midrule[1pt]
\multicolumn{14}{c}{$U_{large}$}                                                                                                                                    \\
\midrule[1pt]
Pseudo Label~\cite{lee2013pseudo}            & 72.41 &	84.06 &	64.54 &	50.05 &	23.62 &	26.80 &	20.13 &	16.66	& 53.25 &	64.69 &	48.52 &	33.47 &	49.76 \tiny{(-2.13)}             \\
Noisy Student~\cite{xie2020self}           & 75.99 &	86.67 &	70.48 &	55.60 &	33.31 &	37.81 &	28.19 &	21.39	& 59.81 &	70.01 &	55.13 &	38.33 &	56.37 \tiny{(+4.48)}               \\
Mean Teacher~\cite{tarvainen2017mean}            & 76.38 &	86.45 &	70.99 &	57.48 &	35.95 &	41.76 &	29.05 &	18.81 &	65.50 &	75.72 &	60.07 &	43.66 &	59.28 \tiny{(+7.39)}                \\
SESS~\cite{zhao2020sess}                    & 75.95 &	86.83 &	70.45 &	55.76 &	34.43 &	40.00 &	27.92 &	19.20 &	63.58 &	74.85 &	58.88 &	39.51 &	57.99 \tiny{(+6.10)}                \\
3DIoUMatch~\cite{wang20203dioumatch}   &  75.81 &	86.11 &	71.82 &	57.84 &	35.70 &	40.68 &	30.34 &	21.15 &	59.69 &	70.69 &	54.92 &	39.08 &	57.07 \tiny{(+5.18)}              \\ 
\bottomrule[2pt]
\end{tabular}}
\vspace{1pt}
\caption{Results of semi-supervised learning methods on the validation split.}
\end{table}

\section{Implementation details}

In this section, we provide implementation and training details for the 3D object detection benchmark.

\subsection{Models for 3D Object Detection}
\textbf{Data split.} We use the training split to train those $6$ models. The performance is evaluated on the validation and testing split.

\textbf{General configurations.} Non Maximum Suppression (NMS) with the IoU threshold $0.01$ is adopted for post-processing. Other configurations are kept the same with the official version of those models if not specially mentioned.

\textbf{Learning scheme.} All the $6$ models are trained with an initial learning rate $0.003$ under the cosine annealing learning scheme. We use the adam optimizer for all the models. The models are trained with the batch size $32$ for $80$ epochs.

\textbf{Data augmentation.} For all the models, we use random flip of the X and Y axis, random rotation from $-45^{\circ}$ to $+45^{\circ}$, random scaling from $0.95$ to $1.05$, and objects cut and paste on the input point cloud as augmentations. We didn't apply augmentations on segmentation maps for PointPainting.

\textbf{PointRCNN.} PointRCNN is a point-based 3D detector that generates proposals directly on point clouds. We sample $60000$ points per frame and construct the segmentation backbone with $32000$-$4000$-$500$-$256$ points. We use the mean size of each category for proposal generation.

\textbf{PointPillars.} PointPillars is a pioneering work that introduces pillar-based representation into 3D object detection. We set the pillar size as $0.2m\times0.2m$ and also use the mean size as the anchor size.

\textbf{SECOND.} SECOND is a voxel-based detector that transforms point clouds into voxels for feature extraction. We set the voxel size as $0.1m\times0.1m\times0.2m$ and use the same anchors as PointPillars.

\textbf{PV-RCNN.} PV-RCNN is a point-voxel based detector that applies SECOND for proposal generation and then utilizes keypoints for RoI feature extraction. We sample $4096$ keypoints per scene.

\textbf{CenterPoints.} CenterPoints introduces center-based target assignments to replace the anchor-based assignments. In addition to the center head, we use the same backbone as SECOND.

\textbf{PointPainting.} PointPainting uses CenterPoints as the 3D detector and HRNet trained on CityScapes to generate semantic segmentation results.

\subsection{Self-Supervised Learning for 3D Object Detection}

\textbf{Data split.} We conduct self-supervised pretraing on the unlabeled sets $U_{\star}$ and then use the training split to finetune models. The performance is evaluated on the validation and testing split.

\textbf{General configurations.} We use the voxel-based SECOND detector as the baseline model for all the methods. During the pretraining stage, we pretrain the backbone of SECOND detector on unlabeled subset. We pretrain those methods for 20 epochs on the $100$k unlabeled subset $U_{small}$, 5 epochs on the $500$k subset $U_{medium}$ and 3 epochs on the $1$ million subset $U_{large}$.

\textbf{Learning scheme.} For all the methods, the pretaining and finetuning learning rate is initialized as $0.003$. We use the adam optimizer and the cosine annealing learning scheme for all the methods.

\textbf{Multi-view augmentation setup}. We generate multi-view of
the original scenes by random flip, scaling with a scale factor sampled
from {[}0.95, 1.05{]} and rotation around vertical yaw axis between
{[}-10, 10{]} degrees. We also do downsampling by a factor sampled
from {[}0.9, 1{]}.

\textbf{PointContrast.} PointContrast defines a contrastive loss over
the point-level features given a pair of overlapping partial scans.
The objective is to minimize the distance between matched points (positive
pairs) and maximize the distance between unmatched ones (negative
pairs). In our setting, we sample a random geometric transformation
to transform an original point cloud scene into $2$ augmented views.
After passing the scenes through SECOND backbone to obtain voxel-wise
features, we randomly select $1024$ voxels within each
scene. The voxel-wise features will be passed through a two-layer
MLP (with dimension $128$, $64$) to project into latent space, with batchnorm and ReLU. The latent space feature will be concatenated with
initial feature and passed through a one-layer MLP with dimension $64$.
The final features will be used for contrastive pretraining. We pretrain
the model using Adam optimizer with the initial learning rate $0.001$ and the batch size as $4$.

\textbf{DeepCluster.} DeepCluster uses k-means clustering to give
each instance a cluster id as the pseudo label and use the label to train the network.
Since clustering method is designed to learn semantic representation,
we randomly crop patches in 3D scenes as pseudo instances and
pass the patches through the backbone to obtain patch-wise features.
We project the features into latent space for clustering and pretraining.
We choose the total cluster number as $100$. Patch-wise
feature will be passed through a two-layer MLP (with dimension $192$,
$128$), with a batchnorm and ReLU layer to project the features to the latent space. This two-layer MLP will not be used in the finetune stage. We pretrain the backbone with Adam optimizer. The initial learning rate
is $0.0048$ with a cosine decay. The batch size is $256$.

\textbf{SwAV.} SwAV improves DeepCluster by introducing prototypes, online clustering and swapped predictions. We use the same clustering and training settings as DeepCluster. For other configurations we follow the settings in the original paper. 

\textbf{BYOL.} BYOL introduces two networks, referred to online network
and target network, that can interact and learn from each other. Given
a 3D scene, we train the online network to predict the target
network's representation of an augmented view of the same scene.
In particular, after passing the 3D scene through the backbone,
we project the representation through a two-layer MLP (with dimension
$4096$, $512$). After that, the predictor in the online
network will project the embedding into a latent space as the final representation of the online network. The predictor is also a two-layer MLP (with dimension
$4096$, $512$). We update the target network by a slow-moving averaging of the online
network with parameter $0.999$. To avoid the model collapsing to trivial
solutions, we further introduce a contrastive regularization term.
Specifically, we follow the design in PointContrast and ramdomly select
some voxel-wise features. A contrastive loss is designed on different
views of the same voxel between its online representation and target
representation. We pretrain the model using Adam optimizer with the initial learning rate $0.001$. 
The batch size is $4$. 


\subsection{Semi-Supervised Learning for 3D Object Detection}

\textbf{Data split.} We first utilize the training split to obtain pretrained teacher and student models, and then we apply semi-supervised learning on the unlabeled set $U_{\star}$. The performance is evaluated on the validation and testing split.

\textbf{General configurations.} We use the SECOND detector as the baseline model for all the methods, which guarantees a fair comparison among those methods. The pretraining process and configurations follow those in C.1.

\textbf{Learning scheme.} The initial learning rate is $0.003$ for both the pretraining and semi-supervised learning process. We use the adam optimizer and the cosine annealing learning scheme for all the methods. For pretraining, the batch size is $32$. For semi-supervised learning, the batch size of labeled data is $8$ and the batch size of unlabeled data is $32$. The pretraining process lasts $80$ epochs. The semi-supervised learning process lasts $25$ epochs for $U_{small}$ and $5$ epochs for $U_{medium}$ and $U_{large}$.


\textbf{Data augmentation.} During the pretraining process, we use random flip of the X and Y axis, random rotation from $-45^{\circ}$ to $+45^{\circ}$, random scaling from $0.95$ to $1.05$, and objects cut and paste on the input point cloud as augmentations. During the semi-supervised learning process, we use the same random flip, random rotation and random scaling for the student model. We didn't apply augmentations on the teacher model.

\textbf{Pseudo Label.} We use the pretrained model to generate pseudo ground truth boxes for each unlabeled scene. The model is then trained with pseudo labels in the unlabeled scenes, as well as real labels in the training split. It is worth noting that we didn't apply any augmentation in the semi-supervised learning process, mainly to explore whether augmentations are necessary with a large amount of data.

\textbf{Mean Teacher.} Mean Teacher uses the teacher and student model for semi-supervised learning. We first load the pretrained weights for both two models, and then the teacher model produces pseudo ground truths to train the student model for the unlabeled subset. A consistency loss is introduced to regularize two models. Specifically, we first match the predicted boxes of student model with pseudo boxes of teacher model by the nearest-neighbor criterion. Then the Kullback–Leibler divergence of class predictions of the matched pairs of boxes is applied as the consistency loss. The teacher model is updated by exponential moving average (EMA) of the student model.

\textbf{Noisy Student.} Noisy Student is a self-training approach in which the student is trained with noise, \ie strong augmentations, using pseudo labels provided by the teacher. After the first round of semi-supervised training, we make the student a new teacher for the second around.

\textbf{SESS.} Self-Ensembling Semi-Supervised (SESS) 3D object detection extends Mean Teacher by introducing another two consistency constraints: size consistency and center consistency, along with class consistency to the matched pairs of boxes. The teacher is also updated by EMA.

\textbf{3DIoUMatch.} 3DIoUMatch introduces an extra IoU prediction head on the detection model. The predicted IoUs are then used for filtering low-quality pseudo boxes. We reject IoU-guided lower-half suppression and the EMA update scheme, since those components are detrimental to the detection performance in our experiments.

\subsection{Unsupervised Domain Adaptation for 3D Object Detection}

\textbf{Data split.} We conduct unsupervised domain adaptation on the training split. The performance is evaluated on the validation split.

\textbf{Learning scheme.} We completely follow the same learning rate, optimizer and learning scheme used in~\cite{yang2021st3d}. 

\textbf{Data augmentation.} We completely follow the augmentations used in~\cite{yang2021st3d}.

\textbf{SN.} Statistical Normalization (SN) is based on the observation that domain gap mainly comes from the differences of object size between different datasets, so this method normalizes the objects' size of the source dataset according to the object statistics on the target domain.

\textbf{ST3D.} ST3D contains two stages: the model is first trained on the source dataset with an augmentation method named random object scaling. Then the model is trained on the target dataset with the aid of pseudo labels and a memory bank.

\section{Visualization} \label{D}

\begin{figure}[H]
\centering
\includegraphics[width=0.8\textwidth]{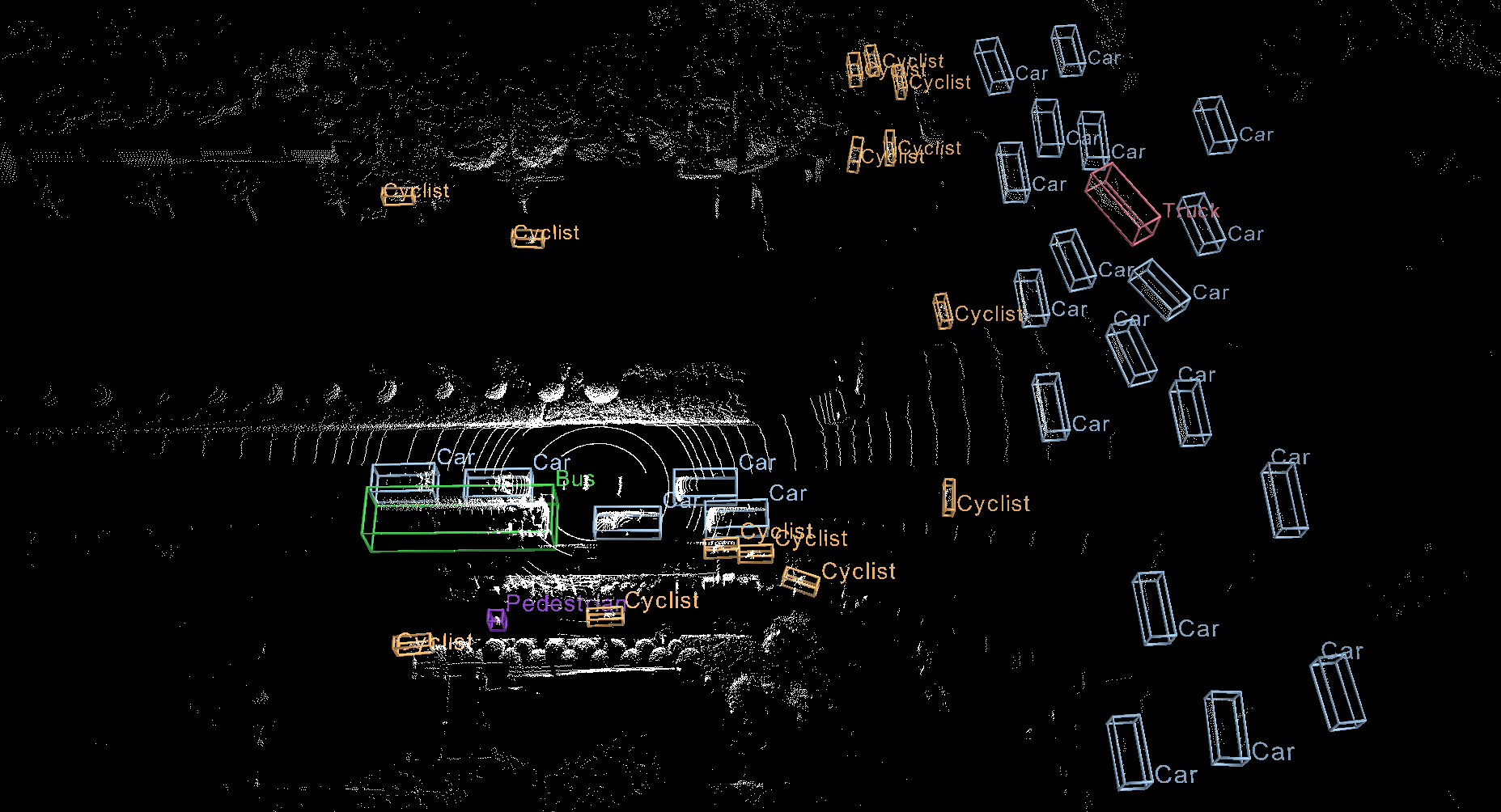}
\caption{Example of 3D annotations.}
\label{fig:example}
\vspace{-6mm}
\end{figure}

\textbf{Annotation example.} We present an example of annotations in Figure~\ref{fig:example}.
\vspace{-4mm}

\begin{figure}[H]
\centering
\includegraphics[width=0.8\textwidth]{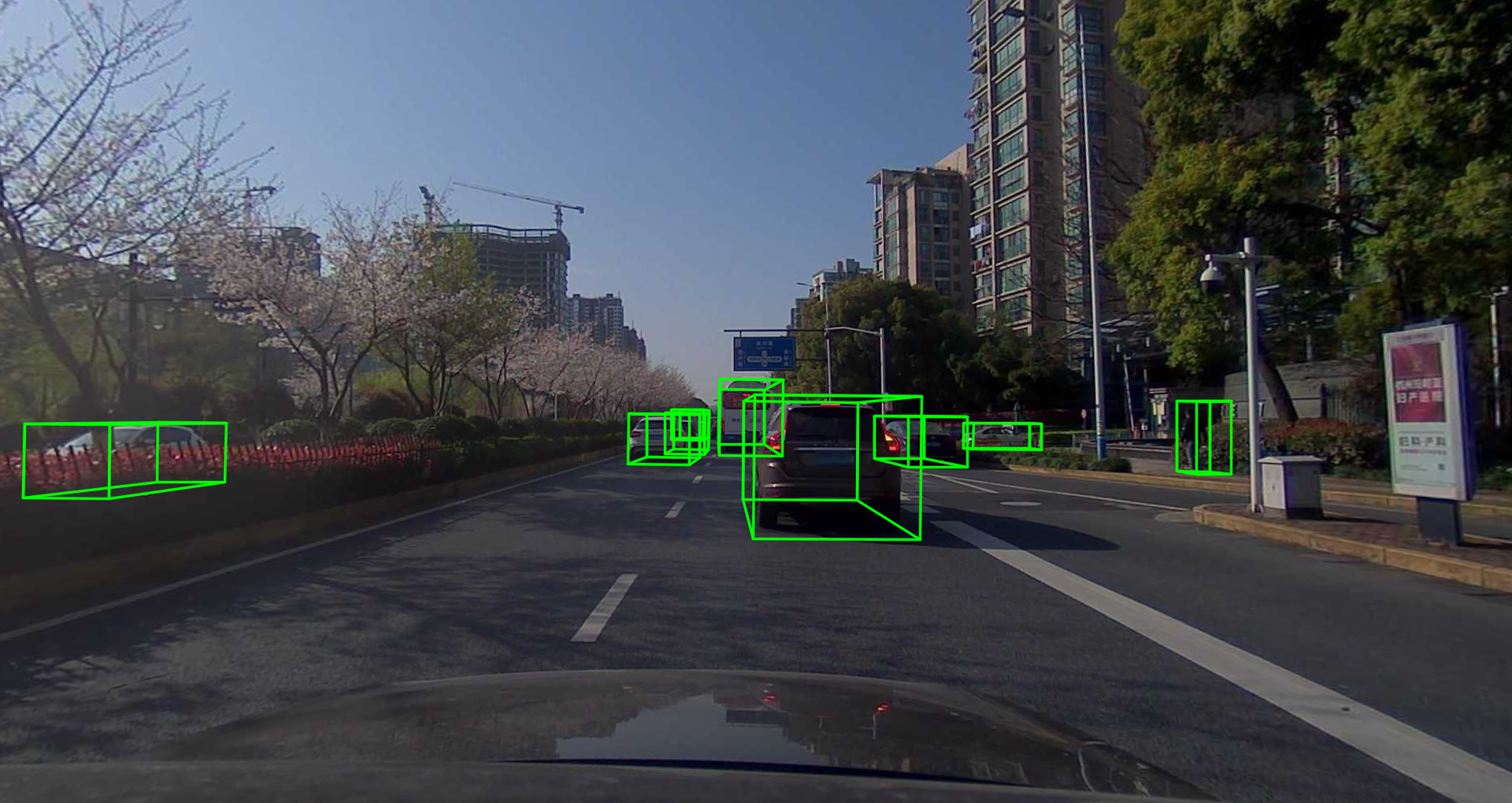}
\caption{Example of 3D annotations on an RGB image.}
\label{fig:proj}
\vspace{-6mm}
\end{figure}

\textbf{3D annotations for RGB images.} We present an example of 3D annotations on an RGB image in Figure~\ref{fig:proj}.  
\vspace{-4mm}

\begin{figure}[H]
\centering
\includegraphics[width=0.8\textwidth]{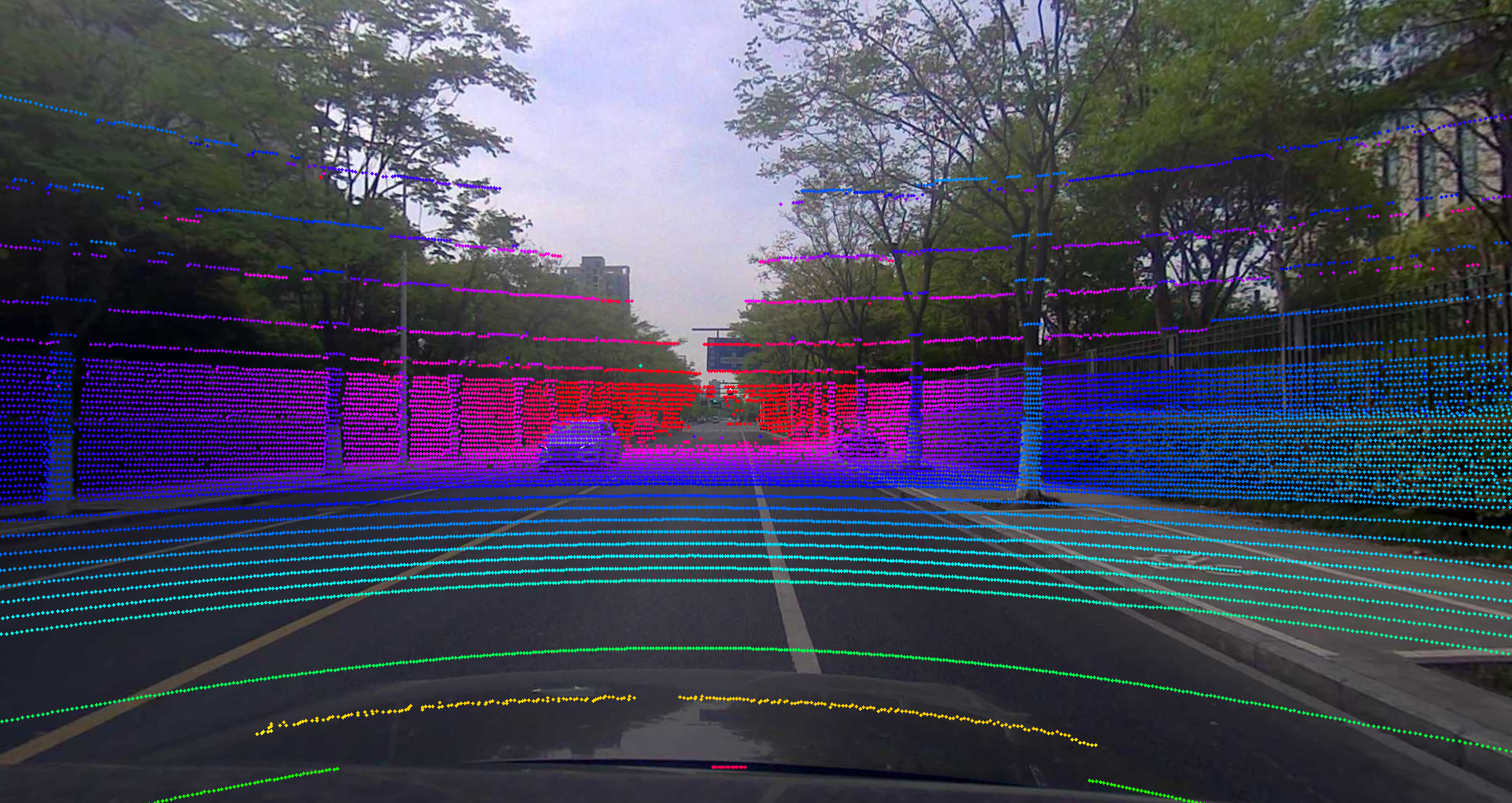}
\caption{Alignment of point cloud and image.}
\label{fig:reflect}
\vspace{-6mm}
\end{figure}

\textbf{Multi-modality alignments.} We present an illustration of the alignments between point clouds and images in Figure~\ref{fig:reflect}. 
\vspace{-3mm}

\begin{figure}[H]
\centering
\includegraphics[width=0.8\textwidth]{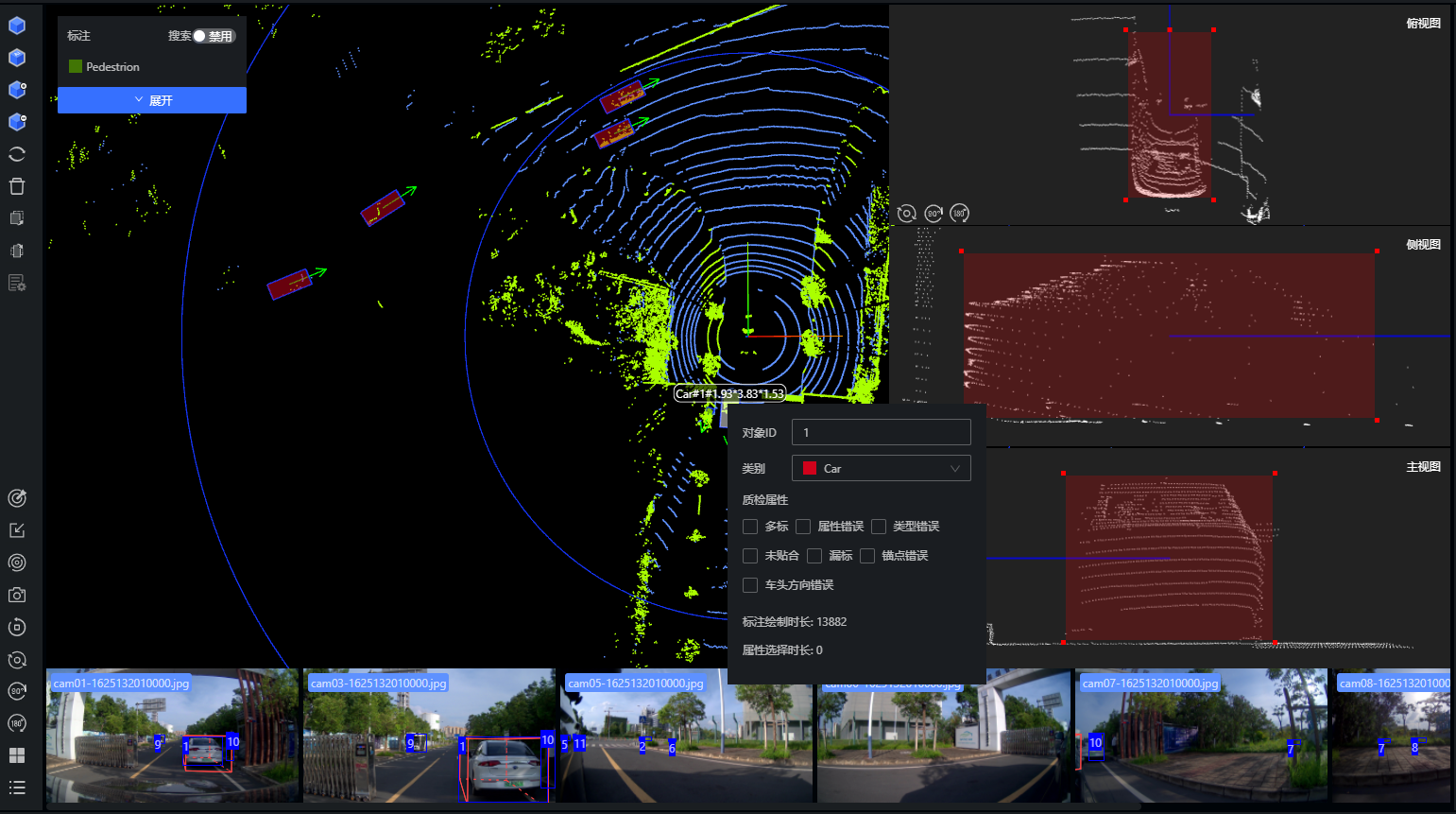}
\caption{Annotation system.}
\label{fig:anno_sys}
\vspace{-6mm}
\end{figure}

\textbf{Annotation system.} We present an interface of the annotation system in Figure~\ref{fig:anno_sys}.
\vspace{-3mm}

\section{Evaluation Metric} \label{E}

Evaluation metric is critical for fair comparisons of different approaches on 3D object detection.
Current 3D IoU-based evaluation metric $AP_{3D}$~\cite{geiger2013vision} faces the problem that objects with opposite orientations can both be matched to the ground truth with the IOU criterion. To resolve this problem, we extend~\cite{geiger2013vision} and take the object orientations into special consideration. In particular, we first re-rank the predictions according to their scores, and set those predicted boxes that have low 3D IoUs with all ground truths of the same category as false positives. The IoU thresholds are $0.7, 0.7, 0.7, 0.3, 0.5$ for car, bus, truck, pedestrian, cyclist respectively. Then we add additional filtering step in which we also set those predictions as false positives if their orientations $\theta$ cannot fall into the $\pm 90^{\circ}$ range of the matched ground truth orientations $\theta^{\prime}$. This step sets a more stringent criterion specially for orientations. The remaining matched predictions are treated as true positives. Finally, we determine $50$ score thresholds with the recall rates $r$ from $0.02$ to $1.00$ at the step $0.02$ and we calculate the corresponding $50$ precision rates to draw the precision-recall curve $p(r)$. The calculation of our orientation-aware $AP^{Ori}_{3D}$ can be formulated as:
\begin{equation}
    AP^{Ori}_{3D} = 100 \int_{0}^{1} max\{p(r^{\prime}|r^{\prime} \ge r)\}dr.
\end{equation}
We merge the car, bus and truck class into a super-class called vehicle following~\cite{sun2020scalability}, so we officially report the $AP^{Ori}_{3D}$ of vehicle, pedestrian and cyclist respectively in the following experiments. We still provide the evaluation interface of $5$ classes for users. Mean AP (mAP) is thus obtained by averaging the scores of $3$ categories. To further inspect the detection performance of different distances, we also provide $AP^{Ori}_{3D}$ of $3$ distance ranges: within $30$m, $30$-$50$m, and farther than $50$m. This is obtained by only considering ground truths and predictions within that distance range. 

\textbf{Discussion on different evaluation metrics.} Current evaluation metrics of 3D detection typically extend the Average Precision (AP) metric~\cite{everingham2010pascal} of 2D detection to the 3D scenarios by changing the matching criterion between the ground truth boxes and predictions. The nuScenes dataset~\cite{caesar2020nuscenes} uses the center distance between boxes on the ground plane as the matching criterion for AP calculation, in ignorance of the size and orientation of the objects. Although the nuScenes detection score (NDS) is proposed to take all factors into consideration, AP still accounts for $50\%$ of the total NDS score, which shows strong preference to the accurate localization of object centers but less attention to the objects' size and orientation. The Waymo Open dataset~\cite{sun2020scalability} applies the Hungarian algorithm to match the ground truths and predictions, which may lead to a higher estimation of AP since objects with no overlaps can also be matched. The KITTI dataset~\cite{geiger2013vision} uses 3D Intersection over Union (IoU) above certain threshold as the matching criterion, but the predicted boxes with the opposite orientations of the ground truths can also be matched, which can be dangerous in practical. In this paper, we extend the 3D IoU-based evaluation metric $AP$ of~\cite{geiger2013vision} and take the object orientations into special consideration. Our orientation-aware $AP^{Ori}_{3D}$ metric is more stringent than~\cite{geiger2013vision}. Compared with the weighted-scoring method in~\cite{sun2020scalability}, our method avoids repeated calculations of the orientation factor, since it has already participated in the computation of 3D rotated IoU. Compared with the distance-based matching scheme~\cite{caesar2020nuscenes}, our method puts equal weights on the object size, center and orientation.

\end{document}